\documentclass[sigconf]{acmart}
\usepackage{booktabs} 
\usepackage{balance}
\usepackage[english]{babel}
\usepackage{moresize}
\usepackage{amsmath}
\usepackage{algorithmic}

\usepackage{comment}
\usepackage{paralist}
\usepackage{bm}
\usepackage{pgfplots}
\usetikzlibrary{pgfplots.dateplot}

\usepackage{flushend}
\usepackage[english]{babel}
\usepackage[latin1]{inputenc}
\usepackage{mathrsfs}
\usepackage{graphicx}

\usepackage{amssymb}
\usepackage{amsfonts}
\usepackage{url}
\usepackage{longtable}
\usepackage{rotating}
\usepackage{multirow}
\usepackage{mathrsfs}
\usepackage{subfigure}
\usepackage{enumitem}
\usepackage[linesnumbered,algoruled,boxed,lined]{algorithm2e}
\usepackage{adjustbox}
\usepackage{hyperref}
\usepackage{pgfplots}
\usetikzlibrary{pgfplots.dateplot}
\usepackage{filecontents}
\definecolor{tblue}{RGB}{31,119,180}
\definecolor{torange}{RGB}{255,127,14}
\definecolor{tgreen}{RGB}{44,160,44}
\definecolor{tred}{RGB}{214,39,40}
\definecolor{tpurple}{RGB}{148,103,189}

\newcommand{\hide}[1]{} 

\newcommand{\ie}{\textit{i}.\textit{e}.}
\newcommand{\eg}{\textit{e}.\textit{g}.} 
\newcommand{\wrt}{\textit{w}.\textit{r}.\textit{t}} 
\newtheorem{Dfn}{Definition}

\def\model{AutoST}

\setcopyright{none}

\copyrightyear{2023}
\acmYear{2023}
\setcopyright{acmlicensed}\acmConference[WWW'23]{Proceedings of the ACM Web Conference 2023}{May 1--5, 2023}{Austin, TX, USA}
\acmBooktitle{Proceedings of the ACM Web Conference 2023 (WWW'23), May 1--5, 2023, Austin, TX, USA}
\acmPrice{15.00}
\acmDOI{10.1145/3543507.3583304}
\acmISBN{978-1-4503-9416-1/23/04}

\begin{document}
\fancyhead{}



\title{Automated Spatio-Temporal Graph Contrastive Learning}

\author{Qianru Zhang}
\affiliation{The University of Hong Kong}
\email{qrzhang@cs.hku.hk}

\author{Chao Huang}
\authornote{Chao Huang is the corresponding author.}
\affiliation{The University of Hong Kong}
\email{chaohuang75@gmail.com}

\author{Lianghao Xia}
\affiliation{The University of Hong Kong}
\email{lhxia@cs.hku.hk}

\author{Zheng Wang}
\affiliation{Huawei Singapore Research Center}
\email{wangzheng155@huawei.com}

\author{Zhonghang Li}
\affiliation{South China University of Technology}
\email{cszhonghang.li@mail.scut.edu.cn}

\author{Siuming Yiu}
\affiliation{The University of Hong Kong}
\email{smyiu@cs.hku.hk}



\begin{abstract}
Among various region embedding methods, graph-based region relation learning models stand out, owing to their strong structure representation ability for encoding spatial correlations with graph neural networks. Despite their effectiveness, several key challenges have not been well addressed in existing methods: i) Data noise and missing are ubiquitous in many spatio-temporal scenarios due to a variety of factors. ii) Input spatio-temporal data (\eg, mobility traces) usually exhibits distribution heterogeneity across space and time. In such cases, current methods are vulnerable to the quality of the generated region graphs, which may lead to suboptimal performance. In this paper, we tackle the above challenges by exploring the \underline{Auto}mated \underline{S}patio-\underline{T}emporal graph contrastive learning paradigm (\model) over the heterogeneous region graph generated from multi-view data sources. Our \model\ framework is built upon a heterogeneous graph neural architecture to capture the multi-view region dependencies with respect to POI semantics, mobility flow patterns and geographical positions. To improve the robustness of our GNN encoder against data noise and distribution issues, we design an automated spatio-temporal augmentation scheme with a parameterized contrastive view generator. \model\ can adapt to the spatio-temporal heterogeneous graph with multi-view semantics well preserved. Extensive experiments for three downstream spatio-temporal mining tasks on several real-world datasets demonstrate the significant performance gain achieved by our \model\ over a variety of baselines. The code is publicly available at \url{https://github.com/HKUDS/AutoST}.
\end{abstract}



\maketitle

\section{Introduction}
\label{sec:intro}

With the advancement of remote sensing technologies and large-scale computing infrastructure, different types of spatio-temporal data are being collected at unprecedented scale from various domains (\eg, intelligent transportation~\cite{trirat2021df}, environmental science~\cite{schweizer2022semi} and public security~\cite{grgic2018human}). Such diverse spatio-temporal data drives the need for effective spatio-temporal prediction frameworks for various urban sensing applications, such as traffic analysis~\cite{wang2020traffic}, human mobility behavior
modeling~\cite{luo2021stan}, and citywide crime prediction~\cite{huang2018deepcrime}. For instance, motivated by the opportunities of building machine learning and big data driven intelligent cities, the discovered human trajectory patterns can help to formulate better urban planning mechanisms~\cite{feng2018deepmove}, or understanding the dynamics of crime occurrences is useful for reducing crime rate~\cite{wang2016crime,li2022spatial}.

To tackle the challenges in urban sensing scenarios, many efforts have been devoted to developing region representation learning techniques, for studying how to learn the complex region dependence structures come in both temporal and spatial dimensions from geographical datasets~\cite{wang2017region}. Instead of the extracting handcrafted region feature design, these region embedding methods enable the automatic discovery of spatial relations among different regions from the collected spatio-temporal data~\cite{yao2018representing,fu2019efficient}. Among various region representation methods, graph-based learning models stand out owing to the strong feature representation effectiveness over the graph-based region relationships~\cite{fu2019efficient, zhang2019unifying,zhang2021multi}. Towards this line, Graph Neural Networks (GNNs), have been utilized as a powerful modeling scheme for aggregating the complex relational information along with the graph connections, such as MV-PN~\cite{fu2019efficient} and MVURE~\cite{zhang2021multi}. Specifically, these methods take inspiration from Graph Auto-Encoder (GAE)~\cite{kipf2016variational} and Graph Attention Network~\cite{velivckovic2017graph}, by following the graph-structured message passing to perform neighborhood feature aggregation. Despite the effectiveness of the existing region representation approaches, we identify two key challenges that have not been well addressed in previous work.\\\vspace{-0.12in}

\noindent \textbf{Data Noise and Incompleteness}. Due to a variety of factors (\eg, high cost of device deployment, or sensor failure), data noise and incompleteness are ubiquitous in the collected spatio-temporal data~\cite{li2021spatial,lin2020preserving}. That is, for graph-based representation methods, the pre-generated spatial graph often contain noisy information with weak dependency between connected regions. For example, a region may not have strong correlations with all its neighboring regions, due to different region functionalities in a city. Therefore, directly aggregating the spatial and temporal information over the generated spatial graph along with the less-relevant region connections will involve the task-irrelevant noisy signals for downstream spatio-temporal mining applications (\eg, crime prediction, traffic analysis).\\\vspace{-0.12in}

\begin{figure}[t]
\centering
\begin{tabular}{c c }
\hspace{-50mm}
  \begin{minipage}{0.23\textwidth}
	\includegraphics[width=\textwidth]{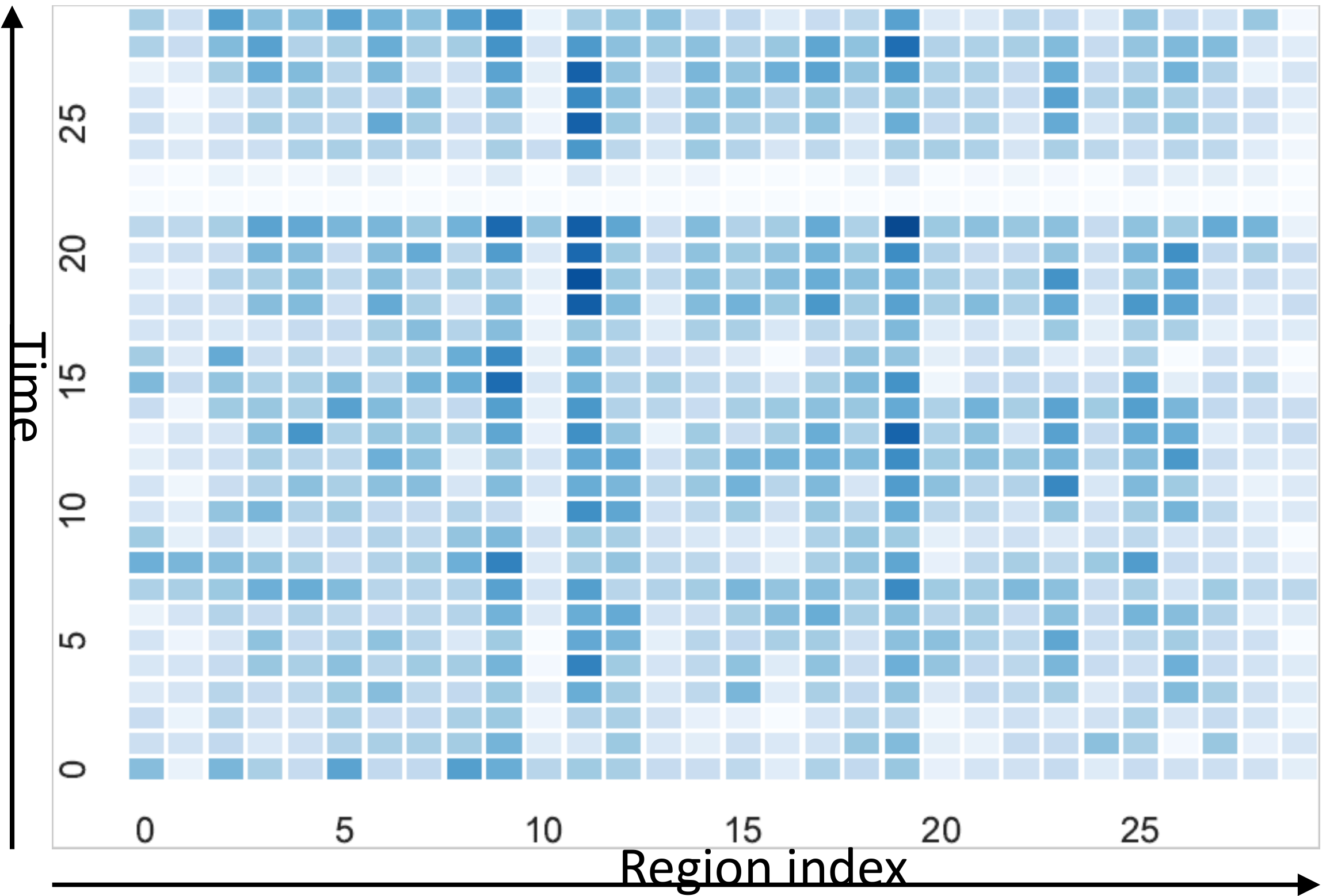}
  \end{minipage}\hspace{-12.0mm}
  &\hspace{-40.0mm}
  \begin{minipage}{0.23\textwidth}
    \includegraphics[width=\textwidth]{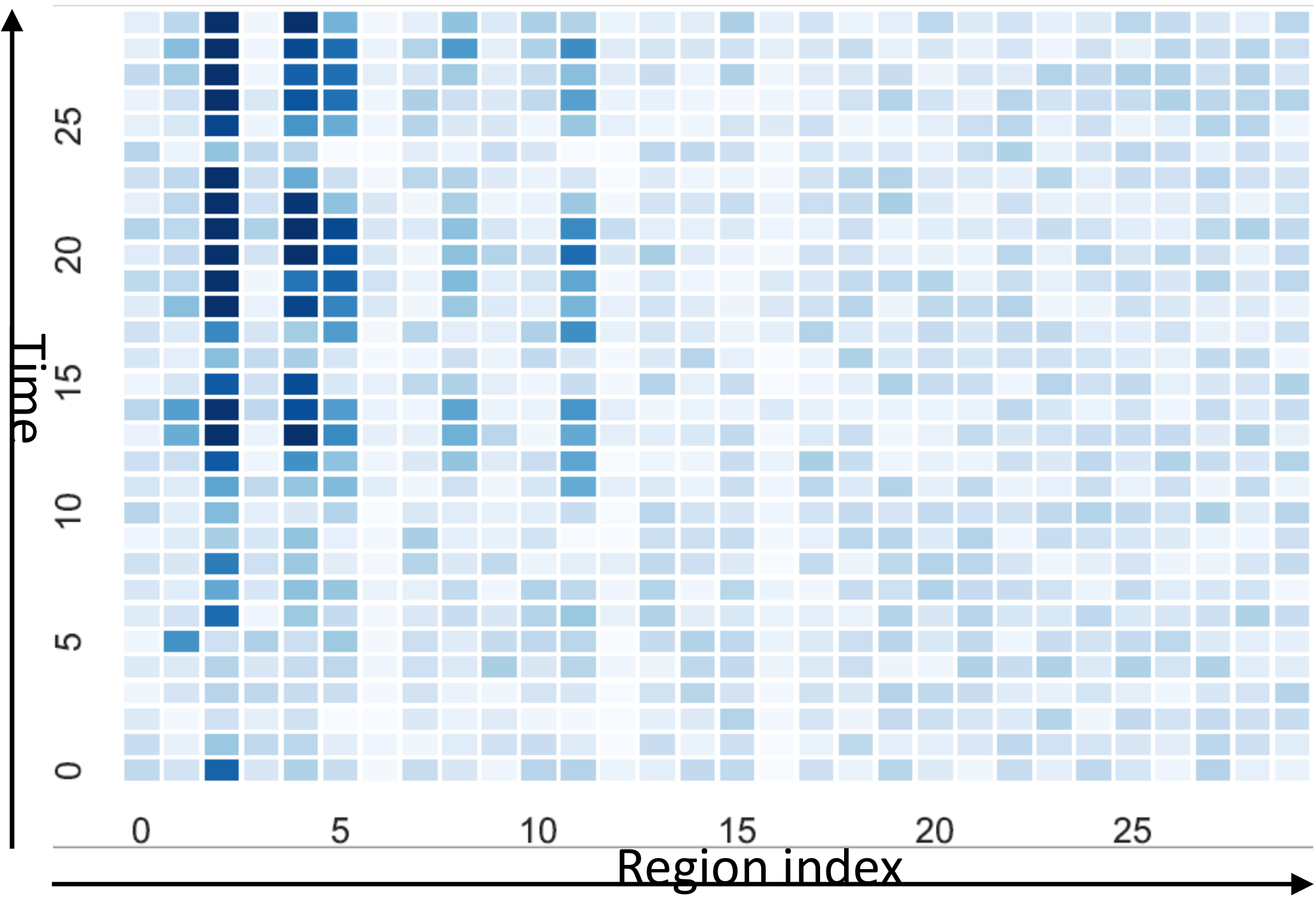}
  \end{minipage}\hspace{-2.0mm}
  \\
\hspace{-0.0mm}
 (a) \small{NYC Trajectory Distribution}
 \hspace{4.0mm}
 (b) \small{CHI Trajectory Distribution}
\end{tabular}
\vspace{-0.1in}
\caption{Trajectory data with spatio-temporal distribution heterogeneity in New York and Chicago on Oct 2016.}
\Description{This Figure illustrates the trajectory distribution heterogeneity of two cities.}
\vspace{-0.15in}
\label{fig:example}
\vspace*{-3mm}
\end{figure}

\noindent \textbf{Spatio-Temporal Distribution Heterogeneity}. spatio-temporal data (\eg, human mobility flow) used in current methods for region graph construction may exhibit distribution heterogeneity across space and time dimension. Such data distribution heterogeneity hinders the representation capability of current neural network models, which leads to the suboptimal region representation performance. To get a better understanding of the skewed data distribution across different regions in urban space, we show the distribution of human mobility trajectories in New York City and Chicago in Figure~\ref{fig:example}. In those figures, the diverse distributions of human mobility data across different geographical regions can be observed, which brings challenges to generate accurate region graph connections to reflect complex spatial correlations based on the mobility traces.

Recently, a series of contrastive-based self-supervised learning (SSL) methods have achieved outstanding performance to address the data noise and label shortage issues, for various tasks in Computer Vision~\cite{simpleCL2020}, Nature Language Processing~\cite{zhang2022frequency}, and Network Embedding~\cite{zhu2021graph}. The core idea of contrastive learning is to generate augmented SSL signals by i) shortening the distance between positive and anchor example in vector representation space; ii) widening the distance between negative and anchor examples~\cite{trivedi2022augmentations,xia2022simgrace}.

Additionally, the spatio-temporal data distributions are also influenced by heterogeneous external factors in urban data scenarios, such as region functionality, environmental attributes, and time-evolving traffic flow. These diverse datasets are exhibited with multi-view in nature. Each data dimension corresponds to view-specific characteristics and semantic. Therefore, how to effectively incorporate such heterogeneous data sources into the region representation framework and fuse knowledge from different views, remains a significant challenge.

In light of these aforementioned motivations and challenges, we propose an \underline{Auto}mated \underline{S}patio-\underline{T}emporal graph contrastive learning (\model) paradigm to provide effective and robust region representations with multi-view spatio-temporal data in urban sensing scenario. Specifically, the spatio-temporal heterogeneous graph neural encoder is firstly introduced to model spatio-temporal patterns by capturing region-wise dependencies from both the intra-view and inter-view data correlations. In our self-supervised learning paradigm, the encoded multi-view region embeddings will serve as the guidance for our augmentation schema to automatically distill the heterogeneous dependencies across all regions in urban space. Our graph contrastive learning component not only enhances the learning process of implicit cross-region relations, but also plays an important role in effectively incorporating auxiliary self-supervised signals in an adaptive manner. Conceptually, our \model\ is advantageous over the existing graph contrastive learning methods in enabling the automatic discovery of useful spatio-temporal SSL signals for more effective data augmentation. Some of current methods directly leverage the random mask operations, which may dropout some useful graph connections and is still vulnerable to noise perturbations.

We highlight the key contributions of this work as follows:
\begin{itemize}[leftmargin=*]

\item \textbf{General Aspect}. We emphasize the importance of jointly tackling the data noise and distribution heterogeneity challenges in spatio-temporal graph learning task. Towards this end, we bring the superiority of contrastive learning into region representation with the automated distilled self-supervised signals, to enhance the robustness and generalization ability of our spatio-temporal heterogeneous graph neural architecture.\\\vspace{-0.12in}

\item \textbf{Methodologies}. We first design spatio-temporal heterogeneous graph neural network to simultaneously capture the intra-view and inter-view region dependencies with the modeling of various spatio-temporal factors. To address the data noise and  distribution diversity issues, we perform the spatio-temporal contrastive learning over the region dependency graph for automated data augmentation. Our automated contrastive learning framework is built upon a variational spatio-temporal graph auto-encoder with our designed parameterized region dependency learner. \\\vspace{-0.12in}

\item \textbf{Experiments Findings}. We validate our \model\ in different spatio-temporal mining tasks, by competing with various region representation and graph neural network baselines.



\end{itemize}



\section{Preliminaries}
\label{sec:model}


In this work, we divide the entire urban area into $I$ (indexed by $i$) geographical regions. To embed each region into latent representation space, we consider three types of spatio-temporal data sources generated from previous $T$ time slots (\eg, days) (indexed by $t$) as model inputs, which are elaborated with the following definitions:\vspace{-0.05in}

\begin{Dfn}
\textbf{Regional Point-of-Interests (POIs) Matrix $\mathcal{P}$}. In urban space, Point-of-Interest information describes the regional functionalities by referring to different categories of geographical places, \eg, restaurant, hotel, medical center and tourist attraction. We define a POI matrix $\mathcal{P} \in\mathbb{R}^{I\times C}$, in which $C$ denotes the number of POI categories. In this work, 50 and 130 POI categories are utilized to generate matrix $\mathcal{P}$ in New York City and Chicago, respectively. Each element $p_{i,c}$ in $\mathcal{P}$ represents the number of places located in the region $r_i$ with the $c$-th category of POIs. Given the generated $\mathcal{P}$, we can associate each region $r_i$ with the POI vector $\mathcal{P}_i \in\mathbb{R}^{1\times C}$. \vspace{-0.05in}
\end{Dfn}

\begin{Dfn}
\textbf{User Mobility Trajectories $\mathcal{M}$}. To reflect the urban dynamics with human mobility data, we collect a set of user mobility trajectories to record human movement. Particularly, each trajectory record is in the format of $(r_s, r_d, t_s, t_d)$, where $r_s$ and $r_d$ denotes the source and destination regions of this trajectory. $t_s$ and $t_d$ indicate the starting and end timestamp of this trajectory, respectively. \vspace{-0.1in}
\end{Dfn}

\begin{Dfn}
\textbf{Region-wise Geographical Distance Matrix $\mathcal{D}$}. We further define a geographical distance matrix $\mathcal{D} \in\mathbb{R}^{I\times I}$ to represent the geographical distance between each pair of regions (\eg, $r_i$ and $r_{i'}$) in the spatial space of a city, based on the centre coordinates (latitude and longitude) of regions. \vspace{-0.05in}
\end{Dfn}

\noindent\textbf{Problem Statement}: Our spatio-temporal graph learning task is: \textbf{Input}: Given the POI matrix $\mathcal{P} \in\mathbb{R}^{I\times C}$, the set of user mobility trajectories $\mathcal{M}$, and the geographical distance matrix $\mathcal{D} \in\mathbb{R}^{I\times I}$. \textbf{Output}: Our goal is to learn a mapping function $f: \bm{r_i} \rightarrow \mathbb{R}^d$ that encodes each region $r_i$ into a low-dimensional embedding with the dimensionality of $d$. The spatial and temporal dependencies between different regions and time slots can be well preserved in the embedding space, benefiting for various urban sensing applications (\eg, traffic forecasting, crime prediction).

\section{Methodology}
\label{sec:solution}

\begin{figure*}
    \centering
    \includegraphics[width=2.1\columnwidth]{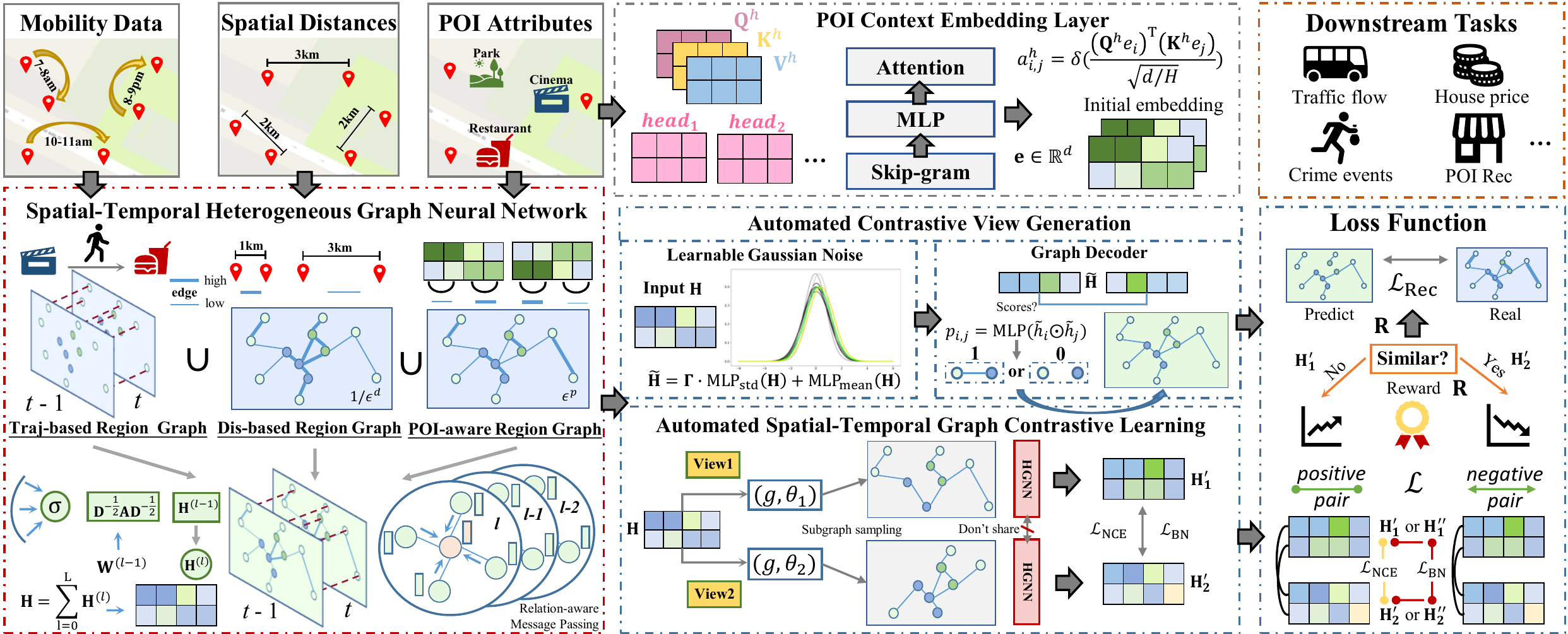}
    \vspace{-0.15in}
    \caption{Architecture of our \model\ region representation model with automated spatio-temporal graph contrastive learning.}
    \Description{This is the framework figure show architecture of our method.}
    \vspace{-0.1in}
    \label{fig:fra_01}
\end{figure*}


In this section, we present the technical details of our \model\ framework. The overall model architecture is illustrated in Figure~\ref{fig:fra_01}.

\vspace{-0.05in}
\subsection{POI Context Embedding Layer}
To encode the Point-of-Interest context into region embeddings, we design a POI context embedding layer to reflect the regional functional information into latent representation space. Motivated by the POI token embedding with Skip-gram~\cite{rahmani2019category}, we feed the region-specific POI vector into the Skip-gram model~\cite{cheng2006n} for POI context embedding. After that, we concatenate the region-specific POI embeddings and generate the POI-aware representations: $\bar{\textbf{E}} = \text{MLP}(\text{ Skip-gram}(\mathcal{P}))$.
\noindent Here, we adopt a Multi-Layer Perceptron to perform the embedding projection. $\textbf{E}\in\mathbb{R}^{I\times d}$ denotes the embedding table for all regions, where $d$ is the hidden dimensionality. The embedding $\textbf{e}_i$ of each region preserves the region-specific POI contextual information into latent semantic space. To endow our POI embedding layer with the modeling of region interactions, we further design the region-wise self-attention mechanism as:
\begin{align}
    \label{eq:selfatt}
    \textbf{e}_i= \mathop{\Bigm|\Bigm|}_{h=1}^H \sum_{j=1}^I \alpha_{i,j}^h \cdot \textbf{V}^{h} \textbf{e}_j; ~~~
    \alpha_{i,j}^h = \delta(\frac{(\textbf{Q}^h\textbf{e}_i)^\top (\textbf{K}^h\textbf{e}_j)}{\sqrt{d/H}})
\end{align}
\noindent where $\textbf{V}^h, \textbf{Q}^h, \textbf{K}^h \in\mathbb{R}^{d/H\times d}$ denote the value, query, and key transformations for the $h$-th attention head, respectively. $H$ denotes the number of heads and $\mathop{\Bigm|\Bigm|}$ denotes concatenating the $H$ vectors. The softmax function $\delta(\cdot)$ is applied. In our self-attention layer, the pairwise interaction between region $r_i$ and $r_{j}$ can be captured in embedding $\textbf{e}_i$ with respect to region-wise POI semantics.

\subsection{Spatio-Temporal Heterogeneous \\ Graph Neural Network}
To comprehensively capture the dynamic region dependencies from different views, \eg, regional functions, traffic flow, and spatial locations, we propose a heterogeneous spatio-temporal GNN. In citywide scenario, our \model\ firstly aims to model the dependencies among different regions in terms of individual representation view. In addition to the importance of cross-region relation modeling, another key dimension is to uncover the inter-correlations among view-specific embeddings of a specific region. For example, implicit dependencies between the POI information and traffic flow of a region are ubiquitous in urban space. Towards this end, in our region graph encoder, both the intra-view and inter-view message passing is performed over the constructed multi-view region graph.


\subsubsection{\bf Heterogeneous Region Graph with Multi-View Relations}
We generate our heterogeneous region graph with multi-typed data views by integrating three view-specific region graphs:\\\vspace{-0.12in}

\begin{itemize}[leftmargin=*]
\item \textbf{POI-aware Region Graph $\mathcal{G}_p$}. We construct a region graph $\mathcal{G}_p$ with the previous encoded Point-of-Interest region embeddings $\textbf{e}_i$ ($1 \leq i \leq I$). To be specific, if the similarity between two region embeddings $\textbf{e}_i$ and $\textbf{e}_{j}$ is larger than the threshold $\epsilon^{p}$, there exists an edge connecting region nodes $r_i$ and $r_{j}$ in graph $\mathcal{G}_p$. Here, the cosine function is applied to measure the embedding similarities. 

\item \noindent\textbf{Trajectory-based Region Graph $\mathcal{G}_m$}. In addition to the stationary POI information, we leverage the time-aware human mobility trajectories to correlate regions, to be reflective of urban flow dynamics over time. We decompose each region $r_i$ into $T$ time slot-specific region nodes ($r_i^t$). In particular, given the record of mobility trajectory $(r_s, r_d, t_s, t_d)$, the source region $r_s^{t_s}$ within the $t_s$-th time slot and destination region $r_d^{t_d}$ within the $t_d$-th time slot will be connected with an edge.

\item \noindent \textbf{Distance-based Region Graph $\mathcal{G}_d$}. To inject the geographical positional information into the graph encoder, we generate a distance-based region graph $\mathcal{G}_d$ by connecting regions in terms of their geographical distance estimated by their coordinates. Specifically, an edge will be added between region $r_i$ and $r_{j}$ if their distance is smaller than the threshold $\epsilon^{d}$.

\end{itemize}

\noindent \textbf{Heterogeneous Region Graph Construction}. Given the above view-specific region graphs $\mathcal{G}_p$, $\mathcal{G}_m$ and $\mathcal{G}_d$, we construct the heterogeneous region graph $\mathcal{G}$ to enable the cross-view region connections and the region temporal self-discrimination connections.

\subsubsection{\bf Relation-aware Message Passing Paradigm}
To capture both the intra-view and inter-view region dependencies based on different types of spatio-temporal data, \model\ conducts message passing over the heterogeneous region graph $\mathcal{G}$ for region embedding. Without loss of generality, we formally present our relation-aware message passing with spatio-temporal connections as follows:
\begin{align}
    \textbf{h}_i^{(l)} = \sigma(\sum_{\gamma \in \Gamma}\sum_{j\in\mathcal{N}^{\gamma}_i} \alpha_{i,\gamma} \textbf{W}_{\gamma}^{(l-1)} \textbf{h}_j^{(l-1)});~~
    \alpha_{i,\gamma} = \frac{1}{|\mathcal{N}^{\gamma}_i|}
\end{align}
\noindent where $\mathcal{N}^{\gamma}_i$ denotes neighbour indices set of region $r_i$ via relation $\gamma$ ($\gamma \in \Gamma$). $\Gamma$ is the relation set. $\textbf{h}_i^{(l)},\textbf{h}_j^{(l-1)}\in\mathbb{R}^d$ represents the embedding vectors for the $i$-th region vertex in the $l$-th graph layer and the $j$-th region vertex in the $(l-1)$-th graph neural layer, respectively. Here, $\textbf{h}_i^{(0)}$ is initialized with $\textbf{e}_i$ which is derived by Eq~\ref{eq:selfatt}. $\sigma(\cdot)$ denotes the ReLU activation function. $\alpha_{i,\gamma}\in\mathbb{R}$ denotes the normalization weight for region vertex pair $(r_i, r_j)$, which is calculated using the degrees of the vertices. 
$\textbf{W}_{\gamma}^{(l-1)}\in\mathbb{R}^{d\times d}$ denotes the learning weights for the $(l-1)-th$ iteration. To fully make use of the information gathered from multi-hop neighborhood, \model\ aggregates the multi-order embeddings together. The cross-view message passing process is given in matrix form shown below:
\begin{align}
    \label{eq:multorder}
    \textbf{H} = \sum_{l=0}^L \textbf{H}^{(l)};~~~
    \textbf{H}^{(l)} = \sigma(\textbf{D}^{-\frac{1}{2}} \textbf{A} \textbf{D}^{-\frac{1}{2}} \textbf{H}^{(l-1)}\textbf{W}_{\gamma}^{(l-1)\top})
\end{align}
where $\textbf{H}\in\mathbb{R}^{|\mathcal{V}|\times d}$ is the embedding matrix, the rows of which are regional embedding vectors $\textbf{h}_i$. $L$ is the number of graph iterations. $\textbf{A}\in\mathbb{R}^{|\mathcal{V}|\times \mathcal{V}|}$ denotes the adjacent matrix with self-loop. $\textbf{D}\in\mathbb{R}^{|\mathcal{V}|\times \mathcal{V}|}$ is the corresponding diagonal degree matrix.

\subsection{Spatio-Temporal GCL}
In this work, \model\ focuses on exploiting the spatio-temporal graph contrastive learning paradigm to tackle the challenges of data noise and distribution heterogeneity. Specifically, our heterogeneous spatio-temporal multi-view graph neural network may be sensitive to the quality of the constructed multi-view region graph $\mathcal{G}$. The message passing scheme may easily aggregate noisy information, which hinders the spatio-temporal representation quality. For example, a region may be connected to another one with close distance but dissimilar region functions, \eg, shopping mall and residential zone. Hence, aggregating the information from irrelevant neighboring nodes in graph $\mathcal{G}$ will impair the region embedding quality, and weaken the representation performance.

Inspired by graph contrastive learning (GCL) in~\cite{you2022bringing}, we propose to generate the contrastive views for spatio-temporal data augmentation in an automatic way. With the designed GCL principle, the graph connection noise issue can be alleviated by incorporating the automated self-supervised signals over the multi-view region graph. In our \model, we develop a variational graph auto-encoder as the graph structure generator for augmentation.

\subsubsection{\bf Variational Graph Encoder}
Inspired by~\cite{you2022bringing}, our \model\ adopts Variational Graph Auto-Encoder (VGAE)~\cite{kipf2016variational} with random walk~\cite{lovasz1993random} for data augmentation, due to the strength of VGAE in considering the data distribution for reconstruction.
To begin with, the VGAE module learns the encoding mapping $\mathcal{G}\rightarrow\mathbb{R}^{|\mathcal{V}|\times d}$ to capture the hierarchical graph structures in low-dimensional node embeddings $\textbf{H}\in\mathbb{R}^{|\mathcal{V}|\times d}$, as defined by Eq~\ref{eq:multorder}. Then, \model\ generates contrastive views in the original graph space by adding Gaussian noises to the low-dimensional graph representations (\ie~the encoded node embeddings), formally as: $\tilde{\textbf{H}} = \mathbf{\Gamma} \cdot \text{MLP}_\text{std}(\textbf{H}) + \text{MLP}_\text{mean}(\textbf{H})$,
\noindent where $\tilde{\textbf{H}}\in\mathbb{R}^{|\mathcal{V}|\times d}$ is the representations for the generated graph view. $\mathbf{\Gamma}\in\mathbb{R}^{|\mathcal{V}|\times d}$ denotes the noise matrix, whose elements $\gamma \sim \text{Gaussian}(\mu, \sigma)$, with $\mu$ and $\sigma$ denoting hyperparameters for the mean and the standard deviation. To realize learnable view generation, \model\ employs two-layer MLPs with trainable parameters (\ie~$\text{MLP}_\text{mean}(\cdot)$ and $\text{MLP}_\text{std}(\cdot)$) to calculate the mean and the standard deviation from $\textbf{H}$.

\subsubsection{\bf Automated Contrastive View Generation}
With the generated representations for graph structures, \model\ then calculates the sampling scores \wrt~the generated view for each node pair, through element-wise product and a MLP. Formally, for node with indices $i$ and $j$, the sampling score $p_{i,j}\in\mathbb{R}$ is calculated by: $p_{i,j} = \text{MLP}(\tilde{\textbf{h}}_i \odot \tilde{\textbf{h}}_j)$,
where $\odot$ denotes the element-wise product. $\tilde{\textbf{h}}_i, \tilde{\textbf{h}}_j\in\mathbb{R}^d$ refer to the $i$-th and the $j$-th rows of the embedding matrix $\tilde{\textbf{H}}$. The probability scores $p_{i,j}$ for all node pairs compose a sampling matrix $\textbf{P}\in\mathbb{R}^{|\mathcal{V}|\times |\mathcal{V}|}$. Next, \model\ sparsify this matrix to acquire a binary adjacent matrix depicting the graph structures of the generated graph view, as $\tilde{\textbf{P}} = \{\tilde{p}_{i,j}\}$, where $\tilde{p}_{i,j} = 0~~\text{if}~p_{i,j}<\epsilon~~\text{else}~~1$.
$\epsilon$ is a hyperparameter for threshold. Here, $\tilde{\textbf{P}}$ is the decoded graph given by our variational graph encoder-decoder. To conduct graph CL efficiently, we apply the random walk algorithm on the decoded graph to generate sub-graphs with less nodes and edges as the contrastive views. Specifically, for dual-view generation, \model\ employs two VGAE modules with the same architecture but non-shared parameters, targeting the two different contrastive views. The decoders generate two graphs, based on which the random walker generates sub-graphs using the same set of seed nodes. Formally, the augmented multi-view graphs are denoted by $\mathcal{G}'_1, \mathcal{G}'_2$ for the two adaptive contrastive representation views. Different from recent graph contrastive learning models with random augmentation operators (\eg, edge perturbation, edge masking, and node dropping~\cite{you2020graph,wu2021self}), we augment the spatio-temporal graph learning with the learnable region dependencies. Therefore, our \model\ enhances the robustness of region representation by adaptively offering auxiliary SSL signals. \\\vspace{-0.12in}

\subsubsection{\bf Contrastive Objective Optimization for Augmentation}
As discussed before, random contrastive view generation methods inevitably lose important structure information and cannot effectively alleviate the side effect caused by noisy and skewed data distribution. To this end, the view generation of our \model\ is augmented with parameterized (\ie, $\text{MLP}_\text{mean}$ and $\text{MLP}_\text{std}$) encoder-decoder architecture and is thus adjustable for contrastive augmentation. To train the learnable VGAE, we calculate the contrastive-based reward for the generated views and combine it with the reconstruction loss of the VGAE. Specifically, the optimization goal for the graph sampling can be formalized as:
\begin{align}
    \label{eq:opt_graphSamp}
    \min\limits_{\theta_1, \theta_2} \text{R}(\mathcal{G}, \theta_1, \theta_2) \cdot \Bigm( \mathcal{L}_\text{Rec}(\mathcal{G}, \theta_1) + \mathcal{L}_\text{Rec}(\mathcal{G}, \theta_2) \Bigm)
\end{align}
\noindent where $\theta_1, \theta_2$ denote the learnable parameters of the two graph samplers for the contrastive views. $\text{R}(\mathcal{G}, \theta_1, \theta_2)$ denotes the reward function based on mutual information maximization and minimization. $\mathcal{L}_\text{Rec}(\cdot)$ denotes the reconstruction loss of VGAE that optimizes the model to rebuild the node-wise connections and the edge-related features from the compressed low-dimensional embeddings $\tilde{\textbf{H}}$.

The principle of assigning reward is to return larger values when the CL loss is big, in which case the view generator has identified challenging contrastive samples. When the CL loss is small, the reward function should return small values to indicate deficiency in the generated views. Instead of using the CL loss as reward directly, \model\ further expands the differences between reward values by minimizing the mutual information between the above two cases. With the Information Minimization (InfoMin)~\cite{tian2020makes}, the reward function is defined as the following discontinuous function:
\begin{align}
\label{eq:infomin_1}
\text{R}_1(\mathcal{G}, \theta_1, \theta_2) = \left\{
    \begin{aligned}
    &1,~~~~~~~\text{if}~ \mathcal{L} (\mathcal{G}, \theta_1, \theta_2)>\epsilon'\\
    &\xi \ll 1~~\text{otherwise}
    \end{aligned}
    \right.
\end{align}
\noindent where $\epsilon'$ denotes the hyperparameter for threshold, and $\xi$ is a pre-selected small value. $\mathcal{L}(\mathcal{G}, \theta_1, \theta_2)$ denotes the contrastive loss function, which will be elaborated later. To enhance our contrastive learning, we further incorporate another dimension of reward signal presented as follows:
\begin{align}
\label{eq:infomin_2}
\text{R}_2(\mathcal{G}, \theta_1, \theta_2) = 
    \begin{aligned}
    1-\cos(\textbf{H}'_1, \textbf{H}'_2)
    \end{aligned}
    .
\end{align}

In concrete, \model\ conducts the heterogeneous graph encoding (Eq~\ref{eq:multorder}) on the sub-graph samples in the two views (\ie~$\mathcal{G}_1', \mathcal{G}_2')$ respectively: $\textbf{H}'_1 = \text{HGNN}(\mathcal{G}'_1)$ and $\textbf{H}'_2 = \text{HGNN}(\mathcal{G}'_2)$
where $\text{HGNN}(\cdot)$ denotes the heterogeneous graph neural network. $\textbf{H}'_1\in\mathbb{R}^{|\mathcal{V}_1'|\times d}, \textbf{H}'_2\in\mathbb{R}^{|\mathcal{V}'_2|\times d}$ represent the resulting node embedding matrices, where $\mathcal{V}_1'$ and $\mathcal{V}_2'$ denote the two generated views, respectively. ${\text{R}(\mathcal{G}, \theta_1, \theta_2)}$ is finally defined: $\text{R}(\mathcal{G}, \theta_1, \theta_2) = w_1 \text{R}_1(\mathcal{G}, \theta_1, \theta_2)+(1-w_1)\text{R}_2(\mathcal{G}, \theta_1, \theta_2)$.
The reconstruction loss $\mathcal{L}_\text{Rec}(\cdot)$ in Eq~\ref{eq:opt_graphSamp} is calculated by minimizing the difference between the reconstructed graph information and the original graph. To be specific, to reconstruct edges, \model\ minimizes the loss: $\mathcal{L}_\text{Rec}(\mathcal{G}, \theta) = -\sum_{(i,j)\in\mathcal{E}} \log(\text{sigm}(p_{i,j})) - \sum_{(i, j)\notin\mathcal{E}} \log(1 - \text{sigm}(p_{i,j}))$.
$\text{sigm}(\cdot)$ denotes the sigmoid function. $p_{i,j}$ is the non-binary sampling score for edge $(i,j)$. 


\subsection{Model Training with Contrastive Learning}
Our \model\ aims to obtain high-quality region embeddings by pretraining the model with self-supervised learning tasks. The optimization goal of \model\ can be formalized as follows:
\begin{align}
    \label{eq:overallLoss}
    \mathcal{L}(\mathcal{G}, \theta_1, \theta_2) = \beta \mathcal{L}_\text{NCE}(\mathcal{G}, \theta_1, \theta_2) \nonumber\\
    + (1 - \beta) \mathcal{L}_\text{BN}(\mathcal{G}, \theta_p, \theta_2)
\end{align}
\noindent where $\mathcal{L}$ denotes the self-supervised loss, which is composed of two terms: the InfoNCE loss~\cite{oord2018representation} $\mathcal{L}_\text{NCE}$, and the Information BottleNeck (InfoBN) $\mathcal{L}_\text{BN}$. The two terms are balanced by the scalar hyperparameter $\beta$. InfoNCE is a commonly-used contrastive learning goal which pulls close the representations for the same entity in two different views, and pushes away the embeddings for different entities in the two views. Based on the embeddings $\textbf{H}'_1$ and $\textbf{H}'_2$, the InfoNCE term is formally defined by:
\begin{equation}
    \label{eq:clLoss}
    \mathcal{L}_\text{NCE} = \sum_{i\in\mathcal{V}_1\cap\mathcal{V}_2} -\log \frac{\exp\Bigm({\cos(\textbf{H}_1'(i), \textbf{H}_2'(i))/\tau}\Bigm)}{ \sum\limits_{j\in\mathcal{V}_1\cap\mathcal{V}_2} \exp\Bigm({\cos(\textbf{H}_1'(i), \textbf{H}_2'(j))/\tau}\Bigm)} \nonumber
\end{equation}
where $\tau$ is the temperature coefficient. $\textbf{H}'(i)\in\mathbb{R}^d$ denotes the embedding vector in the $i$-th row. Here, we use cosine similarity to measure the distance between different representations. Only nodes in both graph views $\mathcal{G}_1$ and $\mathcal{G}_2$ are involved in this loss function.

The self-supervised learning of \model\ is additionally enhanced by the information bottleneck, which further reduce the redundancy in the embeddings from the two views. Specifically, InfoBN generates low-dimensional representations for both views, and minimizes the mutual information between the original contrastive views and their corresponding representations. Formally, InfoBN is to minimize the following loss:
\begin{align}
    \label{eq:bnLoss}
    \mathcal{L}_\text{BN} = -\sum_{i\in\mathcal{V}_1} \log\frac{\exp\Bigm(\cos(\textbf{H}_1'(i), \textbf{H}_1''(i))/\tau \Bigm) }{ \sum\limits_{j\in\mathcal{V}_1} \exp\Bigm(\cos(\textbf{H}_1'(i), \textbf{H}_1''(j))/\tau\Bigm) }\nonumber\\
    -\sum_{i\in\mathcal{V}_2} \log\frac{\exp\Bigm(\cos(\textbf{H}_2'(i), \textbf{H}_2''(i))/\tau\Bigm)}{\sum\limits_{j\in\mathcal{V}_2} \exp\Bigm(\cos(\textbf{H}_2'(i), \textbf{H}_2''(j))/\tau\Bigm) }
\end{align}
\noindent where $\textbf{H}_1''\in\mathbb{R}^{|\mathcal{V}_1|\times d}, \textbf{H}_2''\in\mathbb{R}^{|\mathcal{V}_2|\times d}$ denote the corresponding newly-generated views. This InfoBN regularization diminishes the superfluous information in both views by contrasting them with randomly-augmented representations. 

\section{Evaluation}
\label{sec:eval}


Evaluation is performed to answer the following research questions:
\begin{itemize}[leftmargin=*]

\item \textbf{RQ1.} How does \model\ perform compared with various baselines on different spatio-temporal learning applications?

\item \textbf{RQ2.} How the different data views and contrastive learning components affect the region representation performance?

\item \textbf{RQ3.} How does our \model\ perform in representation learning over regions with different data sparsity degrees?

\item \textbf{RQ4.} What are the benefits of our spatial and temporal dependency modeling across regions with learned representations?\\\vspace{-0.12in}

\item \textbf{RQ5.} How do different settings of hyperparameters affect \model's region representation performance? (Appendix Section)

\end{itemize}

\vspace{-0.1in}
\subsection{Experimental Setup}
\label{sec:setup}
\subsubsection{\bf Datasets and Protocols} We evaluate our \model\ framework on three spatio-temporal mining tasks, \ie, crime prediction, traffic flow forecasting, house price prediction, with several real-world datasets collected from Chicago and New York City. Following the settings in~\cite{xia2021spatial}, different types of crimes are included in Chicago (Theft, Battery, Assault, Damage) and NYC (Burglary, Larceny, Robbery, Assault) datasets. In Appendix Section, we present detailed data description and summarize the data statistics in Table~\ref{fig:data_sta}. We adopt three commonly-used evaluation metrics \emph{MAE}, \emph{MAPE} and \emph{RMSE}, to measure the accuracy in forecasting tasks on urban crimes, citywide traffic flow, and regional house price.

\subsubsection{\bf Baselines for Comparison} are presented as follows:

\noindent \underline{\bf Graph representation methods:}
\begin{itemize}[leftmargin=*]
    \item \textbf{Node2vec~\cite{grover2016node2vec}}. It is a representative network embedding method to encode graph structural information using the random walk-based skip-gram model for embedding nodes.
    \item \textbf{GCN~\cite{kipf2016semi}}. The Graph Convolutional Network is a representative GNN architecture to perform the convolution-based message passing between neighbour nodes.
    \item \textbf{GraphSage~\cite{hamilton2017inductive}}. It enables the information aggregation from the sampled sub-graph structures, so as to improve the computational and memory cost of graph neural networks.
    \item \textbf{GAE~\cite{kipf2016variational}}. Graph Auto-encoder is designed to map nodes into latent embedding space with the input reconstruction objective.
    \item \textbf{GAT~\cite{velivckovic2017graph}}. Graph Attention Network enhances the discrimination ability of graph neural networks by differentiating the relevance degrees among neighboring nodes for propagation.
\end{itemize}

\noindent \underline{\bf SOTA Region representation methods:}
\begin{itemize}[leftmargin=*]
    \item \textbf{POI~\cite{zhang2021multi}}. It is a baseline method which utilizes the POI attributes to represent spatial regions. The TF-IDF algorithm is used to determine the relevance between regions over the POI vectors.
    \item \textbf{HDGE~\cite{wang2017region}}. It is a region representation approach which only relies on the human mobility traces to encode the embedding of each region from the constructed stationary flow graph.
    \item \textbf{ZE-Mob~\cite{yao2018representing}}. This method explores the co-occurrence patterns to model the relations between geographical zones from both the taxi trajectories and human mobility traces.
    \item \textbf{MV-PN~\cite{fu2019efficient}}. It performs representation learning on regions by considering both the POI and human mobility data. The functionality proximities are preserved in region embeddings.
    \item \textbf{CGAL~\cite{zhang2019unifying}}. It is an unsupervised approach to encode region embeddings based on the constructed POI and mobility graphs. The adversarial learning is adopted to integrate the intra-region structures and inter-region dependencies.
    \item \textbf{MVURE~\cite{zhang2021multi}}. This model firstly captures the region correlations based on region geographical properties and user mobility traces.
    \item \textbf{MGFN~\cite{wu2022multi_graph}}. It is a most recently proposed region representation method which constructs the mobility graph to perform the message passing for generating region embeddings.
\end{itemize}

\noindent \underline{\bf Backbone models for crime and traffic prediction.} We select two state-of-the-art methods as the backbone models, to evaluate the quality of the region representations learned by different methods over the tasks of crime prediction and traffic flow forecasting. The encoded region representations are used as the initialized embeddings for the following backbone models.
\begin{itemize}[leftmargin=*]
    \item \textbf{ST-SHN~\cite{xia2021spatial} for Crime Prediction}. It is a state-of-the-art crime prediction method which designs hypergraph structures to capture the global spatial correlations for modeling the crime patterns. Following the settings in the original paper, 128 hyperedges are configured in the hypergraph neural architecture.
    \item \textbf{STGCN~\cite{yu2017spatio} for Traffic Flow Forecasting}. It is a representative traffic prediction model built over a spatio-temporal GCN. The spatio-temporal convolutional blocks jointly encode the dependencies across regions and time slots. Two ST-Conv blocks are adopted to achieve the best performance.
\end{itemize}

\subsubsection{\bf Hyperparameter Settings} Due to space limit, we present the details of parameter settins in \model\ in Appendix Section.

\begin{table*}
\center
\setlength{\abovecaptionskip}{0cm}
\setlength{\belowcaptionskip}{0cm}
\setlength{\tabcolsep}{5pt}
\footnotesize
\caption{Overall performance comparison in urban crime forecasting, traffic prediction, and house price prediction.}
\Description{This table shows whole results of all methods on three tasks in terms of three metrics.}
\label{tab:overall}
\begin{tabular}{|c|c|c|c|c|c|c|c|c|c|c|c|c|c|c|}
    \hline
    \multirow{3}{*}{Model} & \multicolumn{4}{c|}{Crime Prediction} & \multicolumn{6}{c|}{Traffic Prediction} & \multicolumn{4}{c|}{House Price Prediction} \\
    \cline{2-15}
    & \multicolumn{2}{c|}{CHI-Crime} & \multicolumn{2}{c|}{NYC-Crime} & \multicolumn{2}{c|}{CHI-Taxi} & \multicolumn{2}{c|}{NYC-Bike} & \multicolumn{2}{c|}{NYC-Taxi} & \multicolumn{2}{c|}{CHI-House} & \multicolumn{2}{c|}{NYC-House}\\ 
    \cline{2-15}
    & MAE & MAPE & MAE & MAPE & MAE & RMSE & MAE & RMSE & MAE & RMSE & MAE & MAPE & MAE & MAPE\\ \cline{2-15} 
    \hline \hline
    ST-SHN &2.0259 &0.9987 &4.4004 &0.9861 & -- & -- & -- & -- & -- & -- & -- & -- & -- & --  \\
    \hline
    ST-GCN & -- & -- & -- & -- & \multicolumn{1}{c|}{0.1395} &0.5933 & \multicolumn{1}{c|}{0.9240} & 1.8562 & \multicolumn{1}{c|}{1.4093} & 4.1766 & -- & -- & -- & -- \\
    \hline
    Node2vec &1.6334  &0.8605  &4.3646  &0.9454  & \multicolumn{1}{c|}{0.1206} & 0.5803 & \multicolumn{1}{c|}{0.9093} & 1.8513 & \multicolumn{1}{c|}{1.3508} & 4.0105 & \multicolumn{1}{c|}{13137.2178}    &\multicolumn{1}{c|}{44.4278}  & \multicolumn{1}{c|}{4832.6905}    &\multicolumn{1}{c|}{19.8942} \\ 
    \hline
    GCN &1.6061  &0.8546  &4.3257  &0.9234  & \multicolumn{1}{c|}{0.1174} & 0.5707 & \multicolumn{1}{c|}{0.9144} & 1.8321 & \multicolumn{1}{c|}{1.3819} & 4.0200 & \multicolumn{1}{c|}{13074.2121} & \multicolumn{1}{c|}{42.6572} & \multicolumn{1}{c|}{4840.7394} & \multicolumn{1}{c|}{18.3315}\\
    \hline
    GAT &1.5742 &0.8830 &4.3455 &0.9267 & \multicolumn{1}{c|}{0.1105} & 0.5712 & \multicolumn{1}{c|}{0.9110} & 1.8466 & \multicolumn{1}{c|}{1.3746} & 4.0153 & \multicolumn{1}{c|}{13024.7843} &\multicolumn{1}{c|}{43.3221} & \multicolumn{1}{c|}{4799.8482} & 18.3433 \\ 
    \hline
    GraphSage &1.5960 &0.8713 &4.3080 &0.9255 & \multicolumn{1}{c|}{0.1196} & 0.5796 & \multicolumn{1}{c|}{0.9102} & 1.8473 & \multicolumn{1}{c|}{1.3966} & 4.0801 & \multicolumn{1}{c|}{13145.5623} & \multicolumn{1}{c|}{44.3167} & \multicolumn{1}{c|}{4875.6026} & \multicolumn{1}{c|}{18.4570} \\ 
    \hline
    GAE &1.5711 &0.8801 &4.3749 &0.9343 & \multicolumn{1}{c|}{0.1103} & 0.5701 & \multicolumn{1}{c|}{0.9132} & 1.8412 & \multicolumn{1}{c|}{1.3719} & 4.0337 & \multicolumn{1}{c|}{13278.3256} & \multicolumn{1}{c|}{42.3221} & \multicolumn{1}{c|}{4896.9564} & \multicolumn{1}{c|}{18.3114}\\
    \hline
    POI &1.3047 &0.8142 &4.0069 &0.8658 & \multicolumn{1}{c|}{0.0933} & 0.5578 & \multicolumn{1}{c|}{0.8892} & 1.8277 & \multicolumn{1}{c|}{1.3316} & 3.9872 & \multicolumn{1}{c|}{12045.3212} & \multicolumn{1}{c|}{33.5049} & \multicolumn{1}{c|}{4703.3755} & \multicolumn{1}{c|}{16.7920}\\ 
    \hline
    HDGE &1.3586 &0.8273 &4.2021 &0.7821 & \multicolumn{1}{c|}{0.0865} & 0.5502 & \multicolumn{1}{c|}{0.8667} & 1.8251 & \multicolumn{1}{c|}{1.2997} & 3.9846 & \multicolumn{1}{c|}{11976.3215} & \multicolumn{1}{c|}{30.8451} & \multicolumn{1}{c|}{4677.6905} & \multicolumn{1}{c|}{12.5192}\\ 
    \hline
    ZE-Mob &1.3954 &0.8249 &4.3560 &0.8012 & \multicolumn{1}{c|}{0.1002} & 0.5668 & \multicolumn{1}{c|}{0.8900} & 1.8359 & \multicolumn{1}{c|}{1.3314} & 4.0366 & \multicolumn{1}{c|}{12351.1321} & \multicolumn{1}{c|}{38.6171} & \multicolumn{1}{c|}{4730.6927} & \multicolumn{1}{c|}{16.2586}\\ 
    \hline
    MV-PN &1.3370 &0.8132 &4.2342 &0.7791 & \multicolumn{1}{c|}{0.0903} & 0.5502 & \multicolumn{1}{c|}{0.8886} & 1.8313 & \multicolumn{1}{c|}{1.3306} & 3.9530 & \multicolumn{1}{c|}{12565.0607} & \multicolumn{1}{c|}{39.7812} & \multicolumn{1}{c|}{4798.2951} & \multicolumn{1}{c|}{17.0418}\\ 
    \hline
    CGAL &1.3386 &0.7950 &4.1782 &0.7506 & \multicolumn{1}{c|}{0.1013} & 0.5682 & \multicolumn{1}{c|}{0.9097} & 1.8557 & \multicolumn{1}{c|}{1.3353} & 4.0671 & \multicolumn{1}{c|}{12094.5869}    &\multicolumn{1}{c|}{36.9078}     & \multicolumn{1}{c|}{4731.8159}    &\multicolumn{1}{c|}{16.5454}\\
    \hline
    MVURE &1.2586 &0.7087 &3.7683 &0.7318 & \multicolumn{1}{c|}{0.0874} & 0.5405 & \multicolumn{1}{c|}{0.8699} & 1.8157 & \multicolumn{1}{c|}{1.3007} & 3.6715 & \multicolumn{1}{c|}{11095.5323}    &\multicolumn{1}{c|}{34.8954}     & \multicolumn{1}{c|}{4675.1626}    &\multicolumn{1}{c|}{15.9860}\\ 
    \hline
    MGFN &1.2538 &0.6937 &3.5971 &0.7065 & \multicolumn{1}{c|}{0.0831} & 0.5385 & \multicolumn{1}{c|}{0.8783} & 1.8163 & \multicolumn{1}{c|}{1.3266} & 3.7514 & \multicolumn{1}{c|}{10792.7834}    &\multicolumn{1}{c|}{29.9832}    & \multicolumn{1}{c|}{4651.3451}    &\multicolumn{1}{c|}{12.9752}\\ 
    \hline \hline 
    \emph{\model} & \textbf{1.0480} & \textbf{0.4787} & \textbf{2.1073} & \textbf{0.5241} & \textbf{0.0665} & \textbf{0.4931} & \textbf{0.8364} & \textbf{1.8145} & \textbf{1.2871} & \textbf{3.6446} & \textbf{10463.7715} & \textbf{25.7575} & \textbf{4517.7276} & \textbf{7.1660}\\
    \hline
\end{tabular}
\label{tab:over_results}
\vspace{-0.05in}
\end{table*}

\subsection{Effectiveness Evaluation (RQ1)}
\noindent We evaluate our method \model\ on three spatio-temporal mining tasks, including crime prediction, traffic prediction and house price prediction. We provide the result analysis as follows.\\\vspace{-0.1in}

\noindent \textbf{Crime Prediction}. We present the evaluation results of crime prediction on both Chicago and NYC in Table~\ref{tab:over_results} (averaged results) and category-specific results in Table~\ref{fig:crime} (shown in Appendix Section).\vspace{-0.05in}



\begin{itemize}[leftmargin=*]
    \item \model\ achieves the best performance in predicting urban crimes on both datasets, which suggests the effectiveness of our automated spatio-temporal graph contrastive learning paradigm. Specifically, we attribute the improvements to: i) With the design of our hierarchical multi-view graph encoder, \model\ can capture both the intra-view and inter-view region dependencies from multi-typed data sources (\ie, POI semantics, human mobility traces, geographical positions). ii) Benefiting from the adaptive region graph contrastive learning, \model\ offers auxiliary self-supervised signals to enhance the region representation learning against noise perturbation and skewed data distribution.
    
    \item While the current region representation methods (\eg, CGAL, MVURE, MGFN) attempt to capture the region correlations based on the mobility graph, they are vulnerable to the noisy edges in their constructed mobility-based region graphs. The information propagation between less-relevant regions will impair the the performance of region representation learning. In addition, the skewed distribution of mobility data limits the spatial dependency modeling in those baselines. To address these limitations, our \model\ supplements the region embedding paradigm with the region self-discrimination supervision signals, which is complementary to the mobility-based region relationships.
    

    \item The significant performance improvement achieved by \model\ over the compared graph embedding methods (\eg, GCN, GAT, GAE) further confirms the superiority of performing automated data augmentation with the adaptive self-supervision signals. In the graph neural paradigm, the noise effects will be amplified during the message passing over the task-irrelevant region edges. The information aggregated from all the mobility-based or POI-based graph neighboring regions may mislead the encoding of true underlying crime patterns in the urban space.\vspace{-0.05in}

\end{itemize}

\noindent \textbf{Traffic Prediction.}
Table~\ref{tab:over_results} shows the results of predicting traffic volume for future period of 15 mins. (Full version in Table~\ref{fig:traffic}). 


\begin{itemize}[leftmargin=*]
\item \model\ consistently outperforms the compared baselines in all cases, which confirms that our proposed method produces more accurate pre-trained region representations. As the skewed distribution of traffic flow is prevalent across different regions in urban space, with the designed adaptive contrastive learning component over the hierarchical multi-view graph, \model\ overcomes the skewed data distribution issue.
\item Region representation methods (\ie, MVURE, MGFN) require sufficient data to discover traffic patterns or mine human behaviours, which means that they are unable to handle long-tail data. In contrast, contrastive learning with automated view generation has better adaptation on skewed data distribution.

\item Specifically, region representation methods (i.e. MVURE, \model) obtain better performance on long-term traffic prediction (45 minutes) than other traditional methods (i.e. Node2vec, GAE). This suggests region representation methods are effective to capture time-evolving region-wise relations, which are crucial to encode long-term traffic patterns.

\end{itemize}  

\noindent \textbf{Regional House Price Prediction.}
Due to space limit, we summarize the key observations in Section~\ref{sec:house_price} of Appendix.

\vspace{-0.05in}
\subsection{Ablation Study (RQ2)}
In this section, we perform model ablation studies to evaluate the effects of different data views and contrastive learning components of \model\ in contributing the region representation performance. In particular, i) For different data views, we generate two variants: ``w/o $\mathcal{G}_p$'', ``w/o $\mathcal{G}_d$'' by removing POI-aware region graph $\mathcal{G}_p$, and distance-based region graph $\mathcal{G}_d$ respectively. ii) For the ablation study on contrastive learning modules, we generate the variant ``w/o InfoMin'' without the mutual information minimization.



\begin{table}[t]
\center
\setlength{\abovecaptionskip}{-0mm}
\setlength{\belowcaptionskip}{0cm}
\setlength{\tabcolsep}{0.9pt}
\footnotesize
\caption{Ablation Study on Crime and Traffic Prediction}
\Description{This table aims to provide the effect of each component of our method on performance.}
\label{fig:ablation}
\begin{tabular}{|c|cccccccc|}
\hline
\multirow{3}{*}{Model} & \multicolumn{8}{c|}{NYC-Crime}                                                                                                                                                                   \\ \cline{2-9} 
                       & \multicolumn{2}{c|}{Burglary}                        & \multicolumn{2}{c|}{Larceny}                         & \multicolumn{2}{c|}{Robbery}                         & \multicolumn{2}{c|}{Assault}    \\ \cline{2-9} 
                       & \multicolumn{1}{c|}{MAE} & \multicolumn{1}{c|}{MAPE} & \multicolumn{1}{c|}{MAE} & \multicolumn{1}{c|}{MAPE} & \multicolumn{1}{c|}{MAE} & \multicolumn{1}{c|}{MAPE} & \multicolumn{1}{c|}{MAE} & MAPE \\ \hline
w/o $\mathcal{G}_p$                  & \multicolumn{1}{c|}{4.5058}    & \multicolumn{1}{c|}{0.7280}     & \multicolumn{1}{c|}{0.4257}    & \multicolumn{1}{c|}{0.3199}     & \multicolumn{1}{c|}{1.0524}    & \multicolumn{1}{c|}{0.5971}     & \multicolumn{1}{c|}{1.0884}    &0.9654      \\ \hline
w/o $\mathcal{G}_d$                  & \multicolumn{1}{c|}{5.2415}    & \multicolumn{1}{c|}{0.9296}     & \multicolumn{1}{c|}{1.2058}    & \multicolumn{1}{c|}{0.9954}     & \multicolumn{1}{c|}{1.4762}    & \multicolumn{1}{c|}{1.0950}     & \multicolumn{1}{c|}{1.1370}    &0.9813      \\ \hline
w/o InfoMin           & \multicolumn{1}{c|}{4.3684}    & \multicolumn{1}{c|}{0.6666}     & \multicolumn{1}{c|}{0.3773}    & \multicolumn{1}{c|}{0.2967}     & \multicolumn{1}{c|}{0.8918}    & \multicolumn{1}{c|}{0.4227}     & \multicolumn{1}{c|}{0.7584}    &0.6554      \\ \hline
\model                    & \multicolumn{1}{c|}{\textbf{4.2576}}    & \multicolumn{1}{c|}{\textbf{0.6424}}     & \multicolumn{1}{c|}{\textbf{0.3766}}    & \multicolumn{1}{c|}{\textbf{0.2905}}     & \multicolumn{1}{c|}{\textbf{0.8435}}    & \multicolumn{1}{c|}{\textbf{0.3738}}     & \multicolumn{1}{c|}{\textbf{0.7394}}    &\textbf{0.6343}      \\ \hline \hline
\multirow{2}{*}{Model} & \multicolumn{8}{c|}{NYC-Taxi (15/ 30/ 45 min)}                                                                                                                                                       \\ \cline{2-9} 
                       & \multicolumn{4}{c|}{MAE}                                                                                    & \multicolumn{4}{c|}{RMSE}                                                              \\ \hline
w/o $\mathcal{G}_p$                  & \multicolumn{4}{c|}{1.3753/ 1.4470/ 1.5094}                                                                                       & \multicolumn{4}{c|}{3.9952/ 4.8177/ 5.3110}       \\ \hline
w/o $\mathcal{G}_d$                   & \multicolumn{4}{c|}{1.3932/ 1.4505/ 1.5131}                                                                                       & \multicolumn{4}{c|}{4.0802/ 4.9238/ 5.4050}\\ \hline
w/o InfoMin            & \multicolumn{4}{c|}{1.2968/ 1.3622/ 1.3836}                                                                                       & \multicolumn{4}{c|}{3.7405/ 4.7069/ 5.3480} \\ \hline
\model                    & \multicolumn{4}{c|}{\textbf{1.2871/ 1.3134/ 1.3280}}                                                                                       & \multicolumn{4}{c|}{\textbf{3.6446/4.2278/4.6836}}\\ \hline
\end{tabular}
\vspace{-3mm}
\end{table}

\begin{figure*}[t]
    \centering
    \includegraphics[width=1.85\columnwidth]{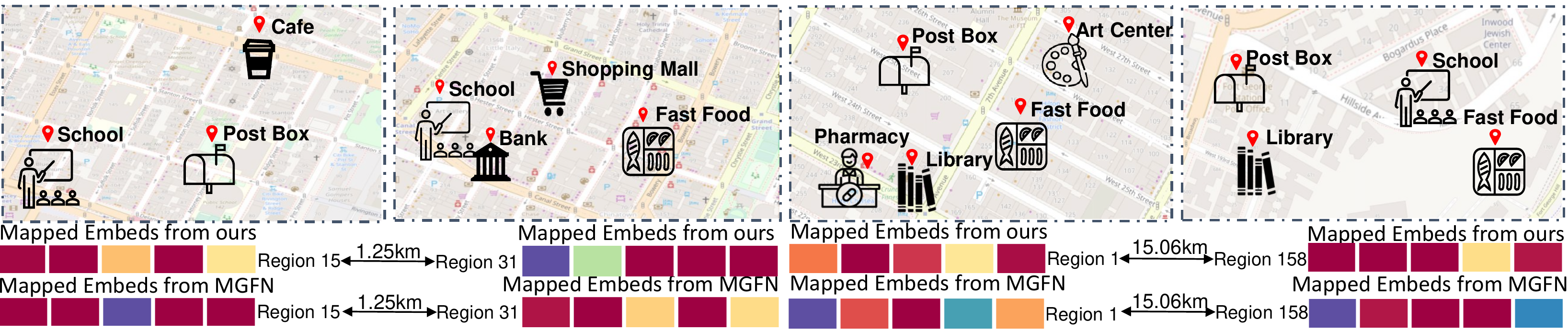}
    \vspace{-0.1in}
    \caption{Case study of our \model\ method on New York City datasets.}
    \Description{The figure is to provide several cases to illustrate the performance of our method on neighbor regions and long distance regions.}
    \label{fig:my_label}
\end{figure*}

\textbf{(1) Analysis on Crime Prediction.} From {Table~\ref{fig:ablation}}, we can observe that the designed component and the incorporated data views in our \model\ framework bring positive effects to the representation performance in the downstream crime prediction. Compared with w/o $\mathcal{G}_p$ and w/o $\mathcal{G}_d$, the performance improvement achieved by \model\ indicates the importance of incorporating region POI semantic and geographical information in region representations. \\\vspace{-0.1in}


\textbf{(2) Analysis on Traffic Prediction.}
Similar results can be observed for traffic prediction task. For example, by incorporating the POI semantic or geographical information into our graph neural encoder, the traffic forecasting performance becomes better with the learned region embeddings by \model. Therefore, with the effectively modeling of heterogeneous dynamic relationships among regions, our method can endow the region representation paradigm with the diverse spatio-temporal patterns.



\vspace{-0.05in}
\subsection{Model Robustness Study (RQ3)}
\begin{figure}
\centering
\begin{tabular}{c c c c}
\hspace{-1.5mm}
\begin{minipage}{0.5cm}
\includegraphics[width=7.5cm]{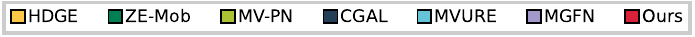}
\end{minipage}\vspace{0.5mm}
&
\\\hspace{-4.0mm}
  \begin{minipage}{0.115\textwidth}
	\includegraphics[width=\textwidth]{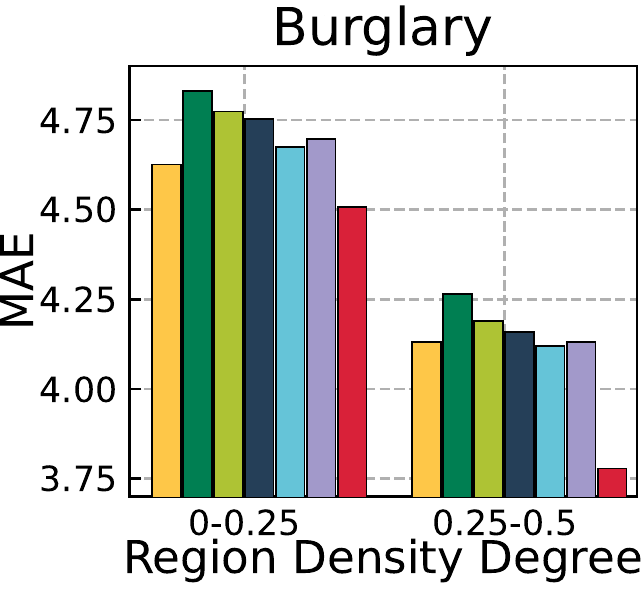}
  \end{minipage}\hspace{-3.5mm}
  &
  \begin{minipage}{0.115\textwidth}
	\includegraphics[width=\textwidth]{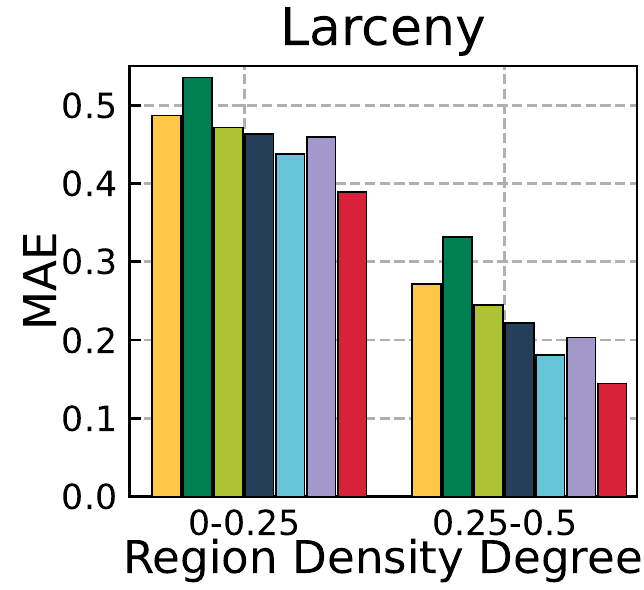}
  \end{minipage}\hspace{-3.5mm}
  &
  \begin{minipage}{0.115\textwidth}
	\includegraphics[width=\textwidth]{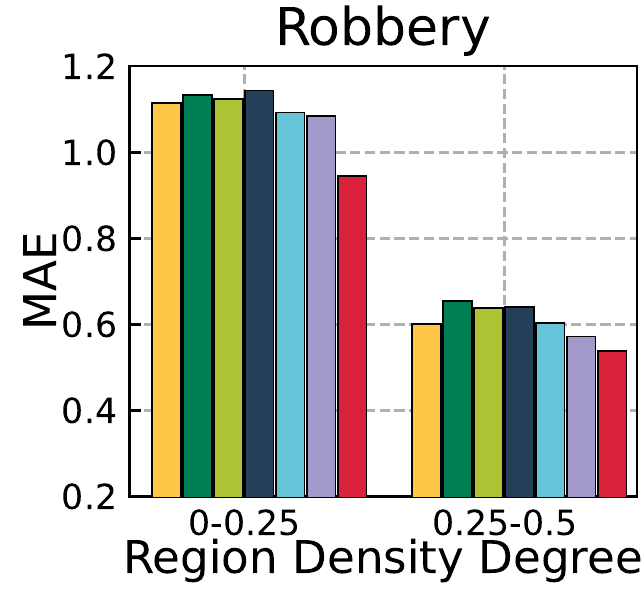}
  \end{minipage}\hspace{-3.5mm}
  &
  \begin{minipage}{0.115\textwidth}
	\includegraphics[width=\textwidth]{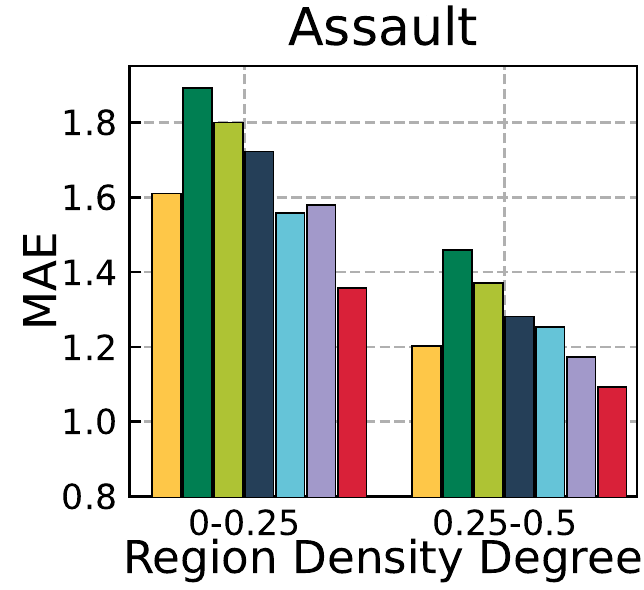}
  \end{minipage}\hspace{-3.5mm}

\end{tabular}
\vspace{-0.1in}
\caption{Results on NYC crime for Burglary, Larceny, Robbery and Assault \wrt\ different data density degrees.}
\Description{The figure is to show the performance of our method on different sparsity degree of data.}
\vspace{-0.15in}
\label{fig:robustness_}
\end{figure}

We also perform experiments to investigate the robustness of our framework \model\ against data sparsity. To achieve this goal, we separately evaluate the prediction accuracy of regions with different density degrees. Here, the density degree of each region is estimated by the ratio of non-zero elements (crime occurs) in the region-specific crime occurrence sequence $\bm Z_r$. Specifically, we partition sparse regions with the crime density degree $\leq$ 0.5 into two groups, \ie, $(0.0, 0.25]$ and $(0.25, 0.5]$. The evaluation results are shown in Figure~\ref{fig:robustness_}. We observe that our \model\ consistently outperforms other methods. As such, this experiment again demonstrates that the spatio-temporal region representation learning benefits greatly from our incorporated self-supervised signals to offer accurate and robust crime prediction on all cases with different crime density degrees. Existing region representation approaches (\eg, MVURE and MGFN) need a large amount of data to mine human behaviours or mobility patterns, which leads lower performance on the sparse data. Another observation admits that current traditional graph learning methods (\eg, GNN-based approaches) can hardly learn high-quality representations on regions with sparse crime data. 



\subsection{Case Study (RQ4)}
In this section, we perform case study to show the capacity of our \model\ in learning the global region dependency with respect to geographical semantics. Specifically, we sample two region pairs, \ie, 1) nearby regions: (region-15, region-31); 2) far away regions: (region-1, region-158). From the presented Figure~\ref{fig:my_label}, while region-15 and region-31 are spatially-adjacent with each other, they exhibit different urban functions. However, their embeddings learned by the baseline MGFN show high similarity. In contrast, the embedding diversity can be observed in the region embeddings learned by our \model\ method. Furthermore, although  region-1 and region-158 are located with long geographical distance, their latent semantics are well preserved in our encoded embddings. Overall, these observations indicate the advantage of our \model\ in capturing global region dependencies in the entire urban space.


\section{Related Work}
\label{sec:relate}

\noindent \textbf{Region Representation Learning}.
Several studies ~\cite{wang2017region,yao2018representing,zhang2021multi,zhang2019unifying,fu2019efficient,wu2022multi_graph} investigate region representation learning problem. In particular,
%
%
%
Fu et al.~\cite{fu2019efficient} propose to improve the quality of region representations by incorporating both intra-region information (\eg, POI distance within a region) and inter-region information (\eg, the similarity of POI distributions between two regions). 
Zhang et al.~\cite{zhang2019unifying} extend the idea proposed in~\cite{fu2019efficient} by introducing a collective adversarial training strategy. 

A recent study proposed in~\cite{zhang2021multi} develops a multi-view joint learning model for learning region representations. It first models the region correlations from different views (\ie, human mobility view and region property view), and then a graph attention mechanism is used for each view in learning region representations.
In addition, Wu et al.~\cite{wu2022multi_graph} propose to extract traffic patterns for learning region representations, but it only considers mobility data and ignores POI data, which is essential for capturing region functionalities. However, the effectiveness of above methods largely relies on generating the high quality region graphs, and may not be able to learn quality region representations under the noisy and skewed-distributed spatio-temporal data. \\\vspace{-0.1in}


\noindent \textbf{Graph Contrastive Learning}. is widely investigated on graph data to learn SSL-enhanced representations ~\cite{you2020graph,sun2019infograph,velickovic2019deep,xia2022hypergraph}. There are two categories of contrastive learning. One is to perform data augmentation with heuristics. For example, some methods~\cite{you2020graph,wu2021self} leverage node/edge mask operations as data augmentor for contrastive self-supervision. However, two issues may exist in those approaches: first, they are specially designed with heuristics, which can hardly generalize to diverse environments. In addition, those solutions select the data augmentation via trial-and-error strategies, which is time-consuming. Inspired by the generative models for data reconstruction, our proposed \model\ performs automatic contrastive learning on spatio-temporal graph so as to distill the informative self-supervised signals for data augmentation.



\vspace{-0.05in}
\section{Conclusion}
\label{sec:conclusoin}

In this paper, we propose a new region representation learning method with the automated spatio-temporal graph contrastive learning paradigm. We explore the adaptive self-supervised learning over the spatio-temporal graphs, and identify the key issues faced by current region representation models. Our work shows that the automated contrastive learning-based data augmentation can offer great potential to region graph representation learning with multi-view spatio-temporal data. We conduct extensive experiments on three spatio-temporal mining tasks with several real-life datasets, to validate the advantage of our proposed \model\ across various settings. In future work, we would like to extend our \model\ to perform the counterfactual learning for distill the causal factors underlying the implicit region-wise correlations.


\section*{Acknowledgments}
This project is partially supported by HKU-SCF FinTech Academy and Shenzhen-Hong Kong-Macao Science and Technology Plan Project (Category C Project: SGDX20210823103537030) and Theme-based Research Scheme T35-710/20-R. This research work is also supported by Department of Computer Science \& Musketeers Foundation Institute of Data Science at the University of Hong Kong.

\clearpage
\bibliographystyle{ACM-Reference-Format}
\balance
\bibliography{sample-base}


\begin{thebibliography}{41}


\ifx \showCODEN    \undefined \def \showCODEN     #1{\unskip}     \fi
\ifx \showDOI      \undefined \def \showDOI       #1{#1}\fi
\ifx \showISBNx    \undefined \def \showISBNx     #1{\unskip}     \fi
\ifx \showISBNxiii \undefined \def \showISBNxiii  #1{\unskip}     \fi
\ifx \showISSN     \undefined \def \showISSN      #1{\unskip}     \fi
\ifx \showLCCN     \undefined \def \showLCCN      #1{\unskip}     \fi
\ifx \shownote     \undefined \def \shownote      #1{#1}          \fi
\ifx \showarticletitle \undefined \def \showarticletitle #1{#1}   \fi
\ifx \showURL      \undefined \def \showURL       {\relax}        \fi
\providecommand\bibfield[2]{#2}
\providecommand\bibinfo[2]{#2}
\providecommand\natexlab[1]{#1}
\providecommand\showeprint[2][]{arXiv:#2}

\bibitem[Zil(2017)]%
        {Zillow.com}
 \bibinfo{year}{2017}\natexlab{}.
\newblock \showarticletitle{Real Estate Value in Chicago}.
\newblock \bibinfo{journal}{\emph{https://www.zillow.com/}}
  (\bibinfo{year}{2017}).
\newblock


\bibitem[Cheng et~al\mbox{.}(2006)]%
        {cheng2006n}
\bibfield{author}{\bibinfo{person}{Winnie Cheng}, \bibinfo{person}{Chris
  Greaves}, {and} \bibinfo{person}{Martin Warren}.}
  \bibinfo{year}{2006}\natexlab{}.
\newblock \showarticletitle{From n-gram to skipgram to concgram}.
\newblock \bibinfo{journal}{\emph{International journal of corpus linguistics}}
  \bibinfo{volume}{11}, \bibinfo{number}{4} (\bibinfo{year}{2006}),
  \bibinfo{pages}{411--433}.
\newblock


\bibitem[Feng et~al\mbox{.}(2018)]%
        {feng2018deepmove}
\bibfield{author}{\bibinfo{person}{Jie Feng}, \bibinfo{person}{Yong Li},
  \bibinfo{person}{Chao Zhang}, \bibinfo{person}{Funing Sun},
  \bibinfo{person}{Fanchao Meng}, \bibinfo{person}{Ang Guo}, {and}
  \bibinfo{person}{Depeng Jin}.} \bibinfo{year}{2018}\natexlab{}.
\newblock \showarticletitle{Deepmove: Predicting human mobility with
  attentional recurrent networks}. In \bibinfo{booktitle}{\emph{International
  Conference on World Wide Web (WWW)}}. \bibinfo{pages}{1459--1468}.
\newblock


\bibitem[Fu et~al\mbox{.}(2019)]%
        {fu2019efficient}
\bibfield{author}{\bibinfo{person}{Yanjie Fu}, \bibinfo{person}{Pengyang Wang},
  \bibinfo{person}{Jiadi Du}, \bibinfo{person}{Le Wu}, {and}
  \bibinfo{person}{Xiaolin Li}.} \bibinfo{year}{2019}\natexlab{}.
\newblock \showarticletitle{Efficient region embedding with multi-view spatial
  networks: A perspective of locality-constrained spatial autocorrelations}. In
  \bibinfo{booktitle}{\emph{AAAI}}, Vol.~\bibinfo{volume}{33}.
  \bibinfo{pages}{906--913}.
\newblock


\bibitem[Grgic-Hlaca et~al\mbox{.}(2018)]%
        {grgic2018human}
\bibfield{author}{\bibinfo{person}{Nina Grgic-Hlaca}, \bibinfo{person}{Elissa~M
  Redmiles}, \bibinfo{person}{Krishna~P Gummadi}, {and} \bibinfo{person}{Adrian
  Weller}.} \bibinfo{year}{2018}\natexlab{}.
\newblock \showarticletitle{Human perceptions of fairness in algorithmic
  decision making: A case study of criminal risk prediction}. In
  \bibinfo{booktitle}{\emph{World Wide Web Conference (WWW)}}.
  \bibinfo{pages}{903--912}.
\newblock


\bibitem[Grover and Leskovec(2016)]%
        {grover2016node2vec}
\bibfield{author}{\bibinfo{person}{Aditya Grover} {and} \bibinfo{person}{Jure
  Leskovec}.} \bibinfo{year}{2016}\natexlab{}.
\newblock \showarticletitle{node2vec: Scalable feature learning for networks}.
  In \bibinfo{booktitle}{\emph{International Conference on Knowledge Discovery
  and Data Mining (KDD)}}. \bibinfo{pages}{855--864}.
\newblock


\bibitem[Hamilton et~al\mbox{.}(2017)]%
        {hamilton2017inductive}
\bibfield{author}{\bibinfo{person}{Will Hamilton}, \bibinfo{person}{Zhitao
  Ying}, {and} \bibinfo{person}{Jure Leskovec}.}
  \bibinfo{year}{2017}\natexlab{}.
\newblock \showarticletitle{Inductive representation learning on large graphs}.
\newblock \bibinfo{journal}{\emph{Advances in Neural Information Processing
  Systems (NIPS)}}  \bibinfo{volume}{30} (\bibinfo{year}{2017}).
\newblock


\bibitem[Huang et~al\mbox{.}(2018)]%
        {huang2018deepcrime}
\bibfield{author}{\bibinfo{person}{Chao Huang}, \bibinfo{person}{Junbo Zhang},
  \bibinfo{person}{Yu Zheng}, {and} \bibinfo{person}{Nitesh~V Chawla}.}
  \bibinfo{year}{2018}\natexlab{}.
\newblock \showarticletitle{DeepCrime: Attentive hierarchical recurrent
  networks for crime prediction}. In \bibinfo{booktitle}{\emph{International
  Conference on Information and Knowledge Management (CIKM)}}.
  \bibinfo{pages}{1423--1432}.
\newblock


\bibitem[Kipf and Welling(2016a)]%
        {kipf2016semi}
\bibfield{author}{\bibinfo{person}{Thomas~N Kipf} {and} \bibinfo{person}{Max
  Welling}.} \bibinfo{year}{2016}\natexlab{a}.
\newblock \showarticletitle{Semi-Supervised Classification with Graph
  Convolutional Networks}. In \bibinfo{booktitle}{\emph{International
  Conference on Learning Representations (ICLR)}}.
\newblock


\bibitem[Kipf and Welling(2016b)]%
        {kipf2016variational}
\bibfield{author}{\bibinfo{person}{Thomas~N Kipf} {and} \bibinfo{person}{Max
  Welling}.} \bibinfo{year}{2016}\natexlab{b}.
\newblock \showarticletitle{Variational graph auto-encoders}.
\newblock \bibinfo{journal}{\emph{arXiv preprint arXiv:1611.07308}}
  (\bibinfo{year}{2016}).
\newblock


\bibitem[Li et~al\mbox{.}(2021)]%
        {li2021spatial}
\bibfield{author}{\bibinfo{person}{Guanyao Li}, \bibinfo{person}{Chih-Chieh
  Hung}, \bibinfo{person}{Mengyun Liu}, \bibinfo{person}{Linfei Pan},
  \bibinfo{person}{Wen-Chih Peng}, {and} \bibinfo{person}{S-H~Gary Chan}.}
  \bibinfo{year}{2021}\natexlab{}.
\newblock \showarticletitle{Spatial-temporal similarity for trajectories with
  location noise and sporadic sampling}. In
  \bibinfo{booktitle}{\emph{International Conference on Data Engineering
  (ICDE)}}. IEEE, \bibinfo{pages}{1224--1235}.
\newblock


\bibitem[Li et~al\mbox{.}(2022)]%
        {li2022spatial}
\bibfield{author}{\bibinfo{person}{Zhonghang Li}, \bibinfo{person}{Chao Huang},
  \bibinfo{person}{Lianghao Xia}, \bibinfo{person}{Yong Xu}, {and}
  \bibinfo{person}{Jian Pei}.} \bibinfo{year}{2022}\natexlab{}.
\newblock \showarticletitle{Spatial-Temporal Hypergraph Self-Supervised
  Learning for Crime Prediction}. In \bibinfo{booktitle}{\emph{International
  Conference on Data Engineering (ICDE)}}. IEEE, \bibinfo{pages}{2984--2996}.
\newblock


\bibitem[Lin et~al\mbox{.}(2020)]%
        {lin2020preserving}
\bibfield{author}{\bibinfo{person}{Haoxing Lin}, \bibinfo{person}{Rufan Bai},
  \bibinfo{person}{Weijia Jia}, \bibinfo{person}{Xinyu Yang}, {and}
  \bibinfo{person}{Yongjian You}.} \bibinfo{year}{2020}\natexlab{}.
\newblock \showarticletitle{Preserving dynamic attention for long-term
  spatial-temporal prediction}. In \bibinfo{booktitle}{\emph{International
  Conference on Knowledge Discovery \& Data Mining (KDD)}}.
  \bibinfo{pages}{36--46}.
\newblock


\bibitem[Lov{\'a}sz(1993)]%
        {lovasz1993random}
\bibfield{author}{\bibinfo{person}{L{\'a}szl{\'o} Lov{\'a}sz}.}
  \bibinfo{year}{1993}\natexlab{}.
\newblock \showarticletitle{Random walks on graphs}.
\newblock \bibinfo{journal}{\emph{Combinatorics, Paul erdos is eighty}}
  \bibinfo{volume}{2}, \bibinfo{number}{1-46} (\bibinfo{year}{1993}),
  \bibinfo{pages}{4}.
\newblock


\bibitem[Luo et~al\mbox{.}(2021)]%
        {luo2021stan}
\bibfield{author}{\bibinfo{person}{Yingtao Luo}, \bibinfo{person}{Qiang Liu},
  {and} \bibinfo{person}{Zhaocheng Liu}.} \bibinfo{year}{2021}\natexlab{}.
\newblock \showarticletitle{Stan: Spatio-temporal attention network for next
  location recommendation}. In \bibinfo{booktitle}{\emph{ACM Web Conference
  (WWW)}}. \bibinfo{pages}{2177--2185}.
\newblock


\bibitem[Oord et~al\mbox{.}(2018)]%
        {oord2018representation}
\bibfield{author}{\bibinfo{person}{Aaron van~den Oord}, \bibinfo{person}{Yazhe
  Li}, {and} \bibinfo{person}{Oriol Vinyals}.} \bibinfo{year}{2018}\natexlab{}.
\newblock \showarticletitle{Representation learning with contrastive predictive
  coding}.
\newblock \bibinfo{journal}{\emph{arXiv preprint arXiv:1807.03748}}
  (\bibinfo{year}{2018}).
\newblock


\bibitem[Rahmani et~al\mbox{.}(2019)]%
        {rahmani2019category}
\bibfield{author}{\bibinfo{person}{Hossein~A Rahmani},
  \bibinfo{person}{Mohammad Aliannejadi}, \bibinfo{person}{Rasoul
  Mirzaei~Zadeh}, \bibinfo{person}{Mitra Baratchi}, \bibinfo{person}{Mohsen
  Afsharchi}, {and} \bibinfo{person}{Fabio Crestani}.}
  \bibinfo{year}{2019}\natexlab{}.
\newblock \showarticletitle{Category-aware location embedding for
  point-of-interest recommendation}. In \bibinfo{booktitle}{\emph{International
  Conference on Research and Development in Information Retrieval (SIGIR)}}.
  \bibinfo{pages}{173--176}.
\newblock


\bibitem[Schweizer et~al\mbox{.}(2022)]%
        {schweizer2022semi}
\bibfield{author}{\bibinfo{person}{Vanessa~Jine Schweizer},
  \bibinfo{person}{Jude~Herijadi Kurniawan}, {and} \bibinfo{person}{Aidan
  Power}.} \bibinfo{year}{2022}\natexlab{}.
\newblock \showarticletitle{Semi-automated Literature Review for Scientific
  Assessment of Socioeconomic Climate Change Scenarios}. In
  \bibinfo{booktitle}{\emph{Companion Proceedings of the Web Conference 2022}}.
  \bibinfo{pages}{789--799}.
\newblock


\bibitem[Sun et~al\mbox{.}(2019)]%
        {sun2019infograph}
\bibfield{author}{\bibinfo{person}{Fan-Yun Sun}, \bibinfo{person}{Jordan
  Hoffmann}, \bibinfo{person}{Vikas Verma}, {and} \bibinfo{person}{Jian Tang}.}
  \bibinfo{year}{2019}\natexlab{}.
\newblock \showarticletitle{Infograph: Unsupervised and semi-supervised
  graph-level representation learning via mutual information maximization}. In
  \bibinfo{booktitle}{\emph{International Conference on Learning
  Representations (ICLR)}}.
\newblock


\bibitem[Tian et~al\mbox{.}(2020)]%
        {tian2020makes}
\bibfield{author}{\bibinfo{person}{Yonglong Tian}, \bibinfo{person}{Chen Sun},
  \bibinfo{person}{Ben Poole}, \bibinfo{person}{Dilip Krishnan},
  \bibinfo{person}{Cordelia Schmid}, {and} \bibinfo{person}{Phillip Isola}.}
  \bibinfo{year}{2020}\natexlab{}.
\newblock \showarticletitle{What makes for good views for contrastive
  learning?}
\newblock \bibinfo{journal}{\emph{Advances in Neural Information Processing
  Systems (NeurIPS)}}  \bibinfo{volume}{33} (\bibinfo{year}{2020}),
  \bibinfo{pages}{6827--6839}.
\newblock


\bibitem[Ting~Chen et~al\mbox{.}(2020)]%
        {simpleCL2020}
\bibfield{author}{\bibinfo{person}{Kevin~Swersky Ting~Chen, Simon~Kornblith}
  {et~al\mbox{.}}} \bibinfo{year}{2020}\natexlab{}.
\newblock \showarticletitle{A simple framework for contrastive learning of
  visual representations}. In \bibinfo{booktitle}{\emph{Advances in Neural
  Information Processing Systems (NIPS)}}.
\newblock


\bibitem[Trirat and Lee(2021)]%
        {trirat2021df}
\bibfield{author}{\bibinfo{person}{Patara Trirat} {and}
  \bibinfo{person}{Jae-Gil Lee}.} \bibinfo{year}{2021}\natexlab{}.
\newblock \showarticletitle{Df-tar: a deep fusion network for citywide traffic
  accident risk prediction with dangerous driving behavior}. In
  \bibinfo{booktitle}{\emph{The Web Conference (WWW)}}.
  \bibinfo{pages}{1146--1156}.
\newblock


\bibitem[Trivedi et~al\mbox{.}(2022)]%
        {trivedi2022augmentations}
\bibfield{author}{\bibinfo{person}{Puja Trivedi}, \bibinfo{person}{Ekdeep~Singh
  Lubana}, \bibinfo{person}{Yujun Yan}, \bibinfo{person}{Yaoqing Yang}, {and}
  \bibinfo{person}{Danai Koutra}.} \bibinfo{year}{2022}\natexlab{}.
\newblock \showarticletitle{Augmentations in graph contrastive learning:
  Current methodological flaws \& towards better practices}. In
  \bibinfo{booktitle}{\emph{ACM Web Conference (WWW)}}.
  \bibinfo{pages}{1538--1549}.
\newblock


\bibitem[Veli{\v{c}}kovi{\'c} et~al\mbox{.}(2018)]%
        {velivckovic2017graph}
\bibfield{author}{\bibinfo{person}{Petar Veli{\v{c}}kovi{\'c}},
  \bibinfo{person}{Guillem Cucurull}, \bibinfo{person}{Arantxa Casanova},
  \bibinfo{person}{Adriana Romero}, \bibinfo{person}{Pietro Lio}, {and}
  \bibinfo{person}{Yoshua Bengio}.} \bibinfo{year}{2018}\natexlab{}.
\newblock \showarticletitle{Graph attention networks}. In
  \bibinfo{booktitle}{\emph{International Conference on Learning
  Representations (ICLR)}}.
\newblock


\bibitem[Velickovic et~al\mbox{.}(2019)]%
        {velickovic2019deep}
\bibfield{author}{\bibinfo{person}{Petar Velickovic}, \bibinfo{person}{William
  Fedus}, \bibinfo{person}{William~L Hamilton}, \bibinfo{person}{Pietro
  Li{\`o}}, \bibinfo{person}{Yoshua Bengio}, {and} \bibinfo{person}{R~Devon
  Hjelm}.} \bibinfo{year}{2019}\natexlab{}.
\newblock \showarticletitle{Deep Graph Infomax.}. In
  \bibinfo{booktitle}{\emph{International Conference on Learning
  Representations (ICLR)}}.
\newblock


\bibitem[Wang et~al\mbox{.}(2016)]%
        {wang2016crime}
\bibfield{author}{\bibinfo{person}{Hongjian Wang}, \bibinfo{person}{Daniel
  Kifer}, \bibinfo{person}{Corina Graif}, {and} \bibinfo{person}{Zhenhui Li}.}
  \bibinfo{year}{2016}\natexlab{}.
\newblock \showarticletitle{Crime rate inference with big data}. In
  \bibinfo{booktitle}{\emph{International Conference on Knowledge Discovery and
  Data Mining (KDD)}}. \bibinfo{pages}{635--644}.
\newblock


\bibitem[Wang and Li(2017)]%
        {wang2017region}
\bibfield{author}{\bibinfo{person}{Hongjian Wang} {and}
  \bibinfo{person}{Zhenhui Li}.} \bibinfo{year}{2017}\natexlab{}.
\newblock \showarticletitle{Region representation learning via mobility flow}.
  In \bibinfo{booktitle}{\emph{International Conference on Information and
  Knowledge Management (CIKM)}}. \bibinfo{pages}{237--246}.
\newblock


\bibitem[Wang et~al\mbox{.}(2020)]%
        {wang2020traffic}
\bibfield{author}{\bibinfo{person}{Xiaoyang Wang}, \bibinfo{person}{Yao Ma},
  \bibinfo{person}{Yiqi Wang}, \bibinfo{person}{Wei Jin}, \bibinfo{person}{Xin
  Wang}, \bibinfo{person}{Jiliang Tang}, \bibinfo{person}{Caiyan Jia}, {and}
  \bibinfo{person}{Jian Yu}.} \bibinfo{year}{2020}\natexlab{}.
\newblock \showarticletitle{Traffic flow prediction via spatial temporal graph
  neural network}. In \bibinfo{booktitle}{\emph{The Web Conference (WWW)}}.
  \bibinfo{pages}{1082--1092}.
\newblock


\bibitem[Wu et~al\mbox{.}(2021)]%
        {wu2021self}
\bibfield{author}{\bibinfo{person}{Lirong Wu}, \bibinfo{person}{Haitao Lin},
  \bibinfo{person}{Cheng Tan}, \bibinfo{person}{Zhangyang Gao}, {and}
  \bibinfo{person}{Stan~Z Li}.} \bibinfo{year}{2021}\natexlab{}.
\newblock \showarticletitle{Self-supervised learning on graphs: Contrastive,
  generative, or predictive}.
\newblock \bibinfo{journal}{\emph{IEEE Transactions on Knowledge and Data
  Engineering (TKDE)}} (\bibinfo{year}{2021}).
\newblock


\bibitem[Wu et~al\mbox{.}(2022)]%
        {wu2022multi_graph}
\bibfield{author}{\bibinfo{person}{Shangbin Wu}, \bibinfo{person}{Xu Yan},
  \bibinfo{person}{Xiaoliang Fan}, \bibinfo{person}{Shirui Pan},
  \bibinfo{person}{Shichao Zhu}, \bibinfo{person}{Chuanpan Zheng},
  \bibinfo{person}{Ming Cheng}, {and} \bibinfo{person}{Cheng Wang}.}
  \bibinfo{year}{2022}\natexlab{}.
\newblock \showarticletitle{Multi-Graph Fusion Networks for Urban Region
  Embedding}.
\newblock \bibinfo{journal}{\emph{arXiv preprint arXiv:2201.09760}}
  (\bibinfo{year}{2022}).
\newblock


\bibitem[Xia et~al\mbox{.}(2022b)]%
        {xia2022simgrace}
\bibfield{author}{\bibinfo{person}{Jun Xia}, \bibinfo{person}{Lirong Wu},
  \bibinfo{person}{Jintao Chen}, \bibinfo{person}{Bozhen Hu}, {and}
  \bibinfo{person}{Stan~Z Li}.} \bibinfo{year}{2022}\natexlab{b}.
\newblock \showarticletitle{SimGRACE: A Simple Framework for Graph Contrastive
  Learning without Data Augmentation}. In \bibinfo{booktitle}{\emph{The Web
  Conference (WWW)}}. \bibinfo{pages}{1070--1079}.
\newblock


\bibitem[Xia et~al\mbox{.}(2021)]%
        {xia2021spatial}
\bibfield{author}{\bibinfo{person}{Lianghao Xia}, \bibinfo{person}{Chao Huang},
  \bibinfo{person}{Yong Xu}, \bibinfo{person}{Peng Dai},
  \bibinfo{person}{Liefeng Bo}, \bibinfo{person}{Xiyue Zhang}, {and}
  \bibinfo{person}{Tianyi Chen}.} \bibinfo{year}{2021}\natexlab{}.
\newblock \showarticletitle{Spatial-Temporal Sequential Hypergraph Network for
  Crime Prediction with Dynamic Multiplex Relation Learning}. In
  \bibinfo{booktitle}{\emph{International Joint Conferences on Artificial
  Intelligence (IJCAI)}}. \bibinfo{pages}{1631--1637}.
\newblock


\bibitem[Xia et~al\mbox{.}(2022a)]%
        {xia2022hypergraph}
\bibfield{author}{\bibinfo{person}{Lianghao Xia}, \bibinfo{person}{Chao Huang},
  \bibinfo{person}{Yong Xu}, \bibinfo{person}{Jiashu Zhao},
  \bibinfo{person}{Dawei Yin}, {and} \bibinfo{person}{Jimmy Huang}.}
  \bibinfo{year}{2022}\natexlab{a}.
\newblock \showarticletitle{Hypergraph contrastive collaborative filtering}. In
  \bibinfo{booktitle}{\emph{International Conference on Research and
  Development in Information Retrieval (SIGIR)}}. \bibinfo{pages}{70--79}.
\newblock


\bibitem[Yao et~al\mbox{.}(2018)]%
        {yao2018representing}
\bibfield{author}{\bibinfo{person}{Zijun Yao}, \bibinfo{person}{Yanjie Fu},
  \bibinfo{person}{Bin Liu}, \bibinfo{person}{Wangsu Hu}, {and}
  \bibinfo{person}{Hui Xiong}.} \bibinfo{year}{2018}\natexlab{}.
\newblock \showarticletitle{Representing urban functions through zone embedding
  with human mobility patterns}. In \bibinfo{booktitle}{\emph{International
  Joint Conferences on Artificial Intelligence (IJCAI)}}.
\newblock


\bibitem[You et~al\mbox{.}(2020)]%
        {you2020graph}
\bibfield{author}{\bibinfo{person}{Yuning You}, \bibinfo{person}{Tianlong
  Chen}, \bibinfo{person}{Yongduo Sui}, \bibinfo{person}{Ting Chen},
  \bibinfo{person}{Zhangyang Wang}, {and} \bibinfo{person}{Yang Shen}.}
  \bibinfo{year}{2020}\natexlab{}.
\newblock \showarticletitle{Graph contrastive learning with augmentations}.
\newblock \bibinfo{journal}{\emph{Advances in Neural Information Processing
  Systems (NIPS)}}  \bibinfo{volume}{33} (\bibinfo{year}{2020}),
  \bibinfo{pages}{5812--5823}.
\newblock


\bibitem[You et~al\mbox{.}(2022)]%
        {you2022bringing}
\bibfield{author}{\bibinfo{person}{Yuning You}, \bibinfo{person}{Tianlong
  Chen}, \bibinfo{person}{Zhangyang Wang}, {and} \bibinfo{person}{Yang Shen}.}
  \bibinfo{year}{2022}\natexlab{}.
\newblock \showarticletitle{Bringing your own view: Graph contrastive learning
  without prefabricated data augmentations}. In
  \bibinfo{booktitle}{\emph{International Conference on Web Search and Data
  Mining (WSDM)}}. \bibinfo{pages}{1300--1309}.
\newblock


\bibitem[Yu et~al\mbox{.}(2017)]%
        {yu2017spatio}
\bibfield{author}{\bibinfo{person}{Bing Yu}, \bibinfo{person}{Haoteng Yin},
  {and} \bibinfo{person}{Zhanxing Zhu}.} \bibinfo{year}{2017}\natexlab{}.
\newblock \showarticletitle{Spatio-temporal graph convolutional networks: A
  deep learning framework for traffic forecasting}. In
  \bibinfo{booktitle}{\emph{International Joint Conferences on Artificial
  Intelligence (IJCAI)}}.
\newblock


\bibitem[Zhang et~al\mbox{.}(2021)]%
        {zhang2021multi}
\bibfield{author}{\bibinfo{person}{Mingyang Zhang}, \bibinfo{person}{Tong Li},
  \bibinfo{person}{Yong Li}, {and} \bibinfo{person}{Pan Hui}.}
  \bibinfo{year}{2021}\natexlab{}.
\newblock \showarticletitle{Multi-view joint graph representation learning for
  urban region embedding}. In \bibinfo{booktitle}{\emph{International Joint
  Conferences on Artificial Intelligence (IJCAI)}}.
  \bibinfo{pages}{4431--4437}.
\newblock


\bibitem[Zhang et~al\mbox{.}(2022)]%
        {zhang2022frequency}
\bibfield{author}{\bibinfo{person}{Tong Zhang}, \bibinfo{person}{Wei Ye},
  \bibinfo{person}{Baosong Yang}, \bibinfo{person}{Long Zhang},
  \bibinfo{person}{Xingzhang Ren}, \bibinfo{person}{Dayiheng Liu},
  \bibinfo{person}{Jinan Sun}, \bibinfo{person}{Shikun Zhang},
  \bibinfo{person}{Haibo Zhang}, {and} \bibinfo{person}{Wen Zhao}.}
  \bibinfo{year}{2022}\natexlab{}.
\newblock \showarticletitle{Frequency-aware contrastive learning for neural
  machine translation}. In \bibinfo{booktitle}{\emph{International Conference
  on Artificial Intelligence (AAAI)}}, Vol.~\bibinfo{volume}{36}.
  \bibinfo{pages}{11712--11720}.
\newblock


\bibitem[Zhang et~al\mbox{.}(2019)]%
        {zhang2019unifying}
\bibfield{author}{\bibinfo{person}{Yunchao Zhang}, \bibinfo{person}{Yanjie Fu},
  \bibinfo{person}{Pengyang Wang}, \bibinfo{person}{Xiaolin Li}, {and}
  \bibinfo{person}{Yu Zheng}.} \bibinfo{year}{2019}\natexlab{}.
\newblock \showarticletitle{Unifying inter-region autocorrelation and
  intra-region structures for spatial embedding via collective adversarial
  learning}. In \bibinfo{booktitle}{\emph{International Conference on Knowledge
  Discovery \& Data Mining (KDD)}}. \bibinfo{pages}{1700--1708}.
\newblock


\bibitem[Zhu et~al\mbox{.}(2021)]%
        {zhu2021graph}
\bibfield{author}{\bibinfo{person}{Yanqiao Zhu}, \bibinfo{person}{Yichen Xu},
  \bibinfo{person}{Feng Yu}, \bibinfo{person}{Qiang Liu}, \bibinfo{person}{Shu
  Wu}, {and} \bibinfo{person}{Liang Wang}.} \bibinfo{year}{2021}\natexlab{}.
\newblock \showarticletitle{Graph contrastive learning with adaptive
  augmentation}. In \bibinfo{booktitle}{\emph{The Web Conference (WWW)}}.
  \bibinfo{pages}{2069--2080}.
\newblock


\end{thebibliography}

\newpage
\appendix \section{Appendix}
\balance
\label{sec:appendix}

In this section, we provide some supplementary materials to support our methodology and evaluation sections with further details. Specifically, we first present the data description and hyperparameter settings. Then, the impacts of key parameters are studied in Section~\ref{sec:parameter_study}. Furthermore, the model convergence analysis of our proposed \model\ and baselines are investigated in Section~\ref{sec:convergence}.

\subsection{Data Description Details}
In our experiments, we partition New York City and Chicago into 180 and 234 disjoint geographical regions based on the census blocks with street boundaries. For the traffic flow datasets, we collect the taxi trips spanning the time period of two weeks from Chicago and New York City to generate the set of user mobility trajectories $\mathcal{M}$. For the collected urban crime datasets, each crime record is formatted as $<\text{crime, category, timestamp, coordinates}>$ and will be mapped into a specific region based on the longitude and latitude information. Following the settings in~\cite{xia2021spatial}, different types of crimes are included in Chicago (Theft, Battery, Assault, Damage) and NYC (Burglary, Larceny, Robbery, Assault) datasets. For the house price dataset, we follow the data pre-processing strategy in~\cite{wang2017region} to generate region-specific price information by considering 22,540 and 44,447 houses~\cite{Zillow.com} in New York City and Chicago, respectively.

\subsection{\bf Hyperparameter Settings}
For fair comparison, the dimensionality $d$ of region representation is set as 96 to be consistent with the settings in~\cite{zhang2021multi,wu2022multi_graph}. The depth of convolutional layers in GCN is set as 3. The learning rate is initialized as 0.0005 with the weight decay of 0.01. For the crime prediction backbone model, ST-SHN~\cite{xia2021spatial} is configured with the learning rate of 0.001 and the weight decay of 0.96. The depth of the spatial path aggregation layers is set as 2. For the traffic prediction backbone model ST-GCN~\cite{yu2017spatio}, the historical time window of all tests are set as 60 minutes with 12 observed data points that are utilized to forecast traffic conditions in the next 15, 30, 45 minutes. Most baselines are implemented with their released codes. \\\vspace{-0.15in}

\subsection{House Price Prediction}
\label{sec:house_price}
The learned region representations with different methods are used as the input embeddings to a Lasso regression method. Table~\ref{tab:over_results} shows the results on the house price data in Chicago and NYC.

\begin{itemize}[leftmargin=*]
    \item We notice that \model\ has the best performance in all cases. Meanwhile, MVURE and MGFN have better performance on Chicago house price data and New York house price data. This is mainly due to they mine human patterns, which build connections between similar or neighbour regions. We also observe that POI method has good performance, since this method builds connections between POIs of regions and central shopping regions including different POIs from that of remote towns. \\\vspace{-0.12in}
    \item Region representations methods achieve better performance than network embeddings methods. The reasons behind this observation is that traditional graph-based methods (\eg, Node2vec, GCN) lack the effective encoding of region-wise dependencies from both spatial and temporal dimensions.
    

\end{itemize}

\begin{table*}
\center
\setlength{\abovecaptionskip}{0cm}
\setlength{\belowcaptionskip}{0cm}
\setlength{\tabcolsep}{3.5pt}

\footnotesize
\caption{Overall performance comparison in crime prediction on both Chicago and NYC datasets.}
\Description{We show the whole evaluation results on each type of crime of two cities.}
\label{fig:crime}
\begin{tabular}{|c|cccccccc|cccccccc|}
\hline
           & \multicolumn{8}{c|}{Chicago}                                                                                                                                                                         & \multicolumn{8}{c|}{New York City}                                                                                                                                                                   \\ \cline{2-17}
Model      & \multicolumn{2}{c|}{Theft}                           & \multicolumn{2}{c|}{Battery}                         & \multicolumn{2}{c|}{Assault}                         & \multicolumn{2}{c|}{Damage}     & \multicolumn{2}{c|}{Burglary}                        & \multicolumn{2}{c|}{Larceny}                         & \multicolumn{2}{c|}{Robbery}                         & \multicolumn{2}{c|}{Assault}    \\ \cline{2-17}
           & \multicolumn{1}{c|}{MAE} & \multicolumn{1}{c|}{MPAE} & \multicolumn{1}{c|}{MAE} & \multicolumn{1}{c|}{MPAE} & \multicolumn{1}{c|}{MAE} & \multicolumn{1}{c|}{MPAE} & \multicolumn{1}{c|}{MAE} & MPAE & \multicolumn{1}{c|}{MAE} & \multicolumn{1}{c|}{MPAE} & \multicolumn{1}{c|}{MAE} & \multicolumn{1}{c|}{MPAE} & \multicolumn{1}{c|}{MAE} & \multicolumn{1}{c|}{MPAE} & \multicolumn{1}{c|}{MAE} & MPAE \\ \hline \hline
ST-SHN & \multicolumn{1}{c|}{1.2100}    & \multicolumn{1}{c|}{0.9995}     & \multicolumn{1}{c|}{1.9275}    & \multicolumn{1}{c|}{0.9993}     & \multicolumn{1}{c|}{2.2018}    & \multicolumn{1}{c|}{0.9998}     & \multicolumn{1}{c|}{2.1349}    &0.9985      & \multicolumn{1}{c|}{5.6100}    & \multicolumn{1}{c|}{0.9975}     & \multicolumn{1}{c|}{1.2000}    & \multicolumn{1}{c|}{0.9987}     & \multicolumn{1}{c|}{1.7551}    & \multicolumn{1}{c|}{0.9998}     & \multicolumn{1}{c|}{1.2909}    &0.9985       \\ \hline
Node2vec   & \multicolumn{1}{c|}{1.1378}    & \multicolumn{1}{c|}{0.9862}     & \multicolumn{1}{c|}{1.7655}    & \multicolumn{1}{c|}{0.8970}     & \multicolumn{1}{c|}{1.9631}    & \multicolumn{1}{c|}{0.9714}     & \multicolumn{1}{c|}{1.9015}    &0.9657      & \multicolumn{1}{c|}{4.9447}    & \multicolumn{1}{c|}{0.8092}     & \multicolumn{1}{c|}{0.7272}    & \multicolumn{1}{c|}{0.6532}     & \multicolumn{1}{c|}{1.0566}    & \multicolumn{1}{c|}{0.8040}     & \multicolumn{1}{c|}{1.2411}    &0.9967      \\ \hline
GCN      & \multicolumn{1}{c|}{1.1065}    & \multicolumn{1}{c|}{0.9643}     & \multicolumn{1}{c|}{1.3012}    & \multicolumn{1}{c|}{0.8094}     & \multicolumn{1}{c|}{1.5431}    & \multicolumn{1}{c|}{0.8094}     & \multicolumn{1}{c|}{1.5031}    &0.8056      & \multicolumn{1}{c|}{4.6993}    & \multicolumn{1}{c|}{0.7912}     & \multicolumn{1}{c|}{0.49994}    & \multicolumn{1}{c|}{0.4178}     & \multicolumn{1}{c|}{1.0655}    & \multicolumn{1}{c|}{0.8004}     & \multicolumn{1}{c|}{1.2407}    &0.9890     \\ \hline
GAT        & \multicolumn{1}{c|}{1.1123}    & \multicolumn{1}{c|}{0.9759}     & \multicolumn{1}{c|}{1.3215}    & \multicolumn{1}{c|}{0.8344}     & \multicolumn{1}{c|}{1.5892}    & \multicolumn{1}{c|}{0.8241}     & \multicolumn{1}{c|}{1.5387}    &0.8277      & \multicolumn{1}{c|}{4.7055}    & \multicolumn{1}{c|}{0.7944}     & \multicolumn{1}{c|}{0.5023}    & \multicolumn{1}{c|}{0.4019}     & \multicolumn{1}{c|}{1.0653}    & \multicolumn{1}{c|}{0.8027}     & \multicolumn{1}{c|}{1.2403}    &0.9949      \\ \hline
GraphSage        & \multicolumn{1}{c|}{1.1231}    & \multicolumn{1}{c|}{0.9790}     & \multicolumn{1}{c|}{1.3574}    & \multicolumn{1}{c|}{0.8561}     & \multicolumn{1}{c|}{1.6016}    & \multicolumn{1}{c|}{0.8563}     & \multicolumn{1}{c|}{1.5761}    &0.8432      & \multicolumn{1}{c|}{4.7313}    & \multicolumn{1}{c|}{0.8066}     & \multicolumn{1}{c|}{0.5213}    & \multicolumn{1}{c|}{0.4314}     & \multicolumn{1}{c|}{1.0719}    & \multicolumn{1}{c|}{0.8110}     & \multicolumn{1}{c|}{1.2418}    &0.9965      \\ \hline
GAE        & \multicolumn{1}{c|}{1.1043}    & \multicolumn{1}{c|}{0.9614}     & \multicolumn{1}{c|}{1.3065}    & \multicolumn{1}{c|}{0.7984}     & \multicolumn{1}{c|}{1.5379}    & \multicolumn{1}{c|}{0.7914}     & \multicolumn{1}{c|}{1.4986}    &0.8033      & \multicolumn{1}{c|}{4.7013}    & \multicolumn{1}{c|}{0.7910}     & \multicolumn{1}{c|}{0.5012}    & \multicolumn{1}{c|}{0.4289}     & \multicolumn{1}{c|}{1.0679}    & \multicolumn{1}{c|}{0.8012}     & \multicolumn{1}{c|}{1.2405}    &0.9958      \\ \hline
POI        & \multicolumn{1}{c|}{0.9733}    & \multicolumn{1}{c|}{0.9341}     & \multicolumn{1}{c|}{1.1065}    & \multicolumn{1}{c|}{0.7513}     & \multicolumn{1}{c|}{1.4089}    & \multicolumn{1}{c|}{0.7541}     & \multicolumn{1}{c|}{1.4076}    &0.7697      & \multicolumn{1}{c|}{4.6939}    & \multicolumn{1}{c|}{0.7825}     & \multicolumn{1}{c|}{0.4969}    & \multicolumn{1}{c|}{0.4172}     & \multicolumn{1}{c|}{1.0660}    & \multicolumn{1}{c|}{0.7970}     & \multicolumn{1}{c|}{1.2400}    &0.9943     \\ \hline
HDGE       & \multicolumn{1}{c|}{0.9545}    & \multicolumn{1}{c|}{0.9012}     & \multicolumn{1}{c|}{1.0887}    & \multicolumn{1}{c|}{0.7389}     & \multicolumn{1}{c|}{1.3970}    & \multicolumn{1}{c|}{0.7217}     & \multicolumn{1}{c|}{1.3768}    &0.7349      & \multicolumn{1}{c|}{4.5658}    & \multicolumn{1}{c|}{0.7160}     & \multicolumn{1}{c|}{0.4734}    & \multicolumn{1}{c|}{0.3930}     & \multicolumn{1}{c|}{1.0507}    & \multicolumn{1}{c|}{0.6731}     & \multicolumn{1}{c|}{1.1551}    &0.9870     \\ \hline
ZE-Mob      & \multicolumn{1}{c|}{1.0983}    & \multicolumn{1}{c|}{0.9547}     & \multicolumn{1}{c|}{1.3142}    & \multicolumn{1}{c|}{0.8236}     & \multicolumn{1}{c|}{1.5345}    & \multicolumn{1}{c|}{0.8163}     & \multicolumn{1}{c|}{1.5138}    &0.8273      & \multicolumn{1}{c|}{4.7570}    & \multicolumn{1}{c|}{0.8013}     & \multicolumn{1}{c|}{0.5186}    & \multicolumn{1}{c|}{0.4307}     & \multicolumn{1}{c|}{1.0722}    & \multicolumn{1}{c|}{0.8093}     & \multicolumn{1}{c|}{1.1403}    &0.9975      \\ \hline
MV-PN     & \multicolumn{1}{c|}{0.9613}    & \multicolumn{1}{c|}{0.9146}     & \multicolumn{1}{c|}{1.0946}    & \multicolumn{1}{c|}{0.7452}     & \multicolumn{1}{c|}{1.4013}    & \multicolumn{1}{c|}{0.7393}     & \multicolumn{1}{c|}{1.3582}    &0.7218      & \multicolumn{1}{c|}{4.6329}    & \multicolumn{1}{c|}{0.7502}     & \multicolumn{1}{c|}{0.4213}    & \multicolumn{1}{c|}{0.3708}     & \multicolumn{1}{c|}{1.0642}    & \multicolumn{1}{c|}{0.7840}     & \multicolumn{1}{c|}{1.1091}    &0.9982      \\ \hline
CGAL      & \multicolumn{1}{c|}{0.9589}    & \multicolumn{1}{c|}{0.9014}     & \multicolumn{1}{c|}{1.0897}    & \multicolumn{1}{c|}{0.7403}     & \multicolumn{1}{c|}{1.3995}    & \multicolumn{1}{c|}{0.7345}     & \multicolumn{1}{c|}{1.3698}    &0.7296      & \multicolumn{1}{c|}{4.6013}    & \multicolumn{1}{c|}{0.7203}     & \multicolumn{1}{c|}{0.4113}    & \multicolumn{1}{c|}{0.3651}     & \multicolumn{1}{c|}{1.0714}    & \multicolumn{1}{c|}{0.7765}     & \multicolumn{1}{c|}{1.1009}    &0.9894      \\ \hline
MVURE      & \multicolumn{1}{c|}{0.9365}    & \multicolumn{1}{c|}{0.8910}     & \multicolumn{1}{c|}{1.0631}    & \multicolumn{1}{c|}{0.6957}     & \multicolumn{1}{c|}{1.3709}    & \multicolumn{1}{c|}{0.6375}     & \multicolumn{1}{c|}{1.3037}    &0.6567      & \multicolumn{1}{c|}{4.5907}    & \multicolumn{1}{c|}{0.7144}     & \multicolumn{1}{c|}{0.4077}    & \multicolumn{1}{c|}{0.3262}     & \multicolumn{1}{c|}{1.0578}    & \multicolumn{1}{c|}{0.5889}     & \multicolumn{1}{c|}{0.8410}    &0.6943     \\ \hline
MGFN      & \multicolumn{1}{c|}{0.9231}    & \multicolumn{1}{c|}{0.9015}     & \multicolumn{1}{c|}{1.0804}    & \multicolumn{1}{c|}{0.5824}     & \multicolumn{1}{c|}{1.3016}    & \multicolumn{1}{c|}{0.6072}     & \multicolumn{1}{c|}{1.2563}    &0.6503      & \multicolumn{1}{c|}{4.5646}    & \multicolumn{1}{c|}{0.7994}     & \multicolumn{1}{c|}{0.4285}    & \multicolumn{1}{c|}{0.3084}     & \multicolumn{1}{c|}{1.0475}    & \multicolumn{1}{c|}{0.6310}     & \multicolumn{1}{c|}{0.8319}    &0.7096     \\ \hline \hline 
\model  & \multicolumn{1}{c|}{\textbf{0.9075}}    & \multicolumn{1}{c|}{\textbf{0.8711}}     & \multicolumn{1}{c|}{\textbf{0.8915}}    & \multicolumn{1}{c|}{\textbf{0.5710}}     & \multicolumn{1}{c|}{\textbf{1.1296}}    & \multicolumn{1}{c|}{\textbf{0.4254}}     & \multicolumn{1}{c|}{\textbf{0.9158}}    &\textbf{0.4542}      & \multicolumn{1}{c|}{\textbf{4.2576}}    & \multicolumn{1}{c|}{\textbf{0.6424}}     & \multicolumn{1}{c|}{\textbf{0.3766}}    & \multicolumn{1}{c|}{\textbf{0.2905}}     & \multicolumn{1}{c|}{\textbf{0.8435}}    & \multicolumn{1}{c|}{\textbf{0.3738}}     & \multicolumn{1}{c|}{\textbf{0.7394}}    &\textbf{0.6343}      \\ \hline
\end{tabular}
\end{table*}

\begin{table*}
\center
\setlength{\abovecaptionskip}{0cm}
\setlength{\belowcaptionskip}{0cm}
\setlength{\tabcolsep}{2.5pt}
\footnotesize
\caption{Overall performance comparison in traffic prediction with different periods of time.}
\Description{This table present whole traffic prediction results of three time steps.}
\label{fig:traffic}
\begin{tabular}{|c|cc|cc|cc|}
\hline
\multirow{2}{*}{Model} & \multicolumn{2}{c|}{CHI-Taxi (15/ 30/ 45 min)}            & \multicolumn{2}{c|}{NYC-Bike (15/ 30/ 45 min)}       & \multicolumn{2}{c|}{NYC-Taxi (15/ 30/ 45 min)}       \\ \cline{2-7} 
                       & \multicolumn{1}{c|}{MAE} & RMSE & \multicolumn{1}{c|}{MAE} & RMSE & \multicolumn{1}{c|}{MAE} & RMSE \\ \hline \hline
STGCN                   & \multicolumn{1}{c|}{0.1395/ 0.2051/ 0.2385}    &0.5933/ 0.6198/ 0.6330           & \multicolumn{1}{c|}{0.9240/ 1.0601/ 1.2984}    &1.8562/ 2.1823/ 2.8658          & \multicolumn{1}{c|}{1.4093/ 1.4513/ 1.5159}    &4.1766/ 4.7814/ 5.4858         \\ \hline
Node2vec               & \multicolumn{1}{c|}{0.1206/ 0.1407/ 0.1694}    &0.5803/ 0.6054/ 0.63327          & \multicolumn{1}{c|}{0.9093/ 0.9875/ 1.1512}    &1.8513/ 2.0917/ 2.8019         & \multicolumn{1}{c|}{1.3508/ 1.4078/ 1.4637}    & 4.0105/ 5.0608/ 5.4043         \\ \hline
GCN                    & \multicolumn{1}{c|}{0.1174/ 0.1368/ 0.1639}    &0.5707/ 0.6016/ 0.6278          & \multicolumn{1}{c|}{0.9144/ 1.0238/ 1.2175}    &1.8321/ 2.1036/ 2.8359          & \multicolumn{1}{c|}{1.3819/ 1.4309/ 1.4934}    & 4.0200/ 4.8711/ 5.3463         \\ \hline
GAT                    & \multicolumn{1}{c|}{0.1105/ 0.1317/ 0.1619}    &0.5712/ 0.6028/ 0.6305          & \multicolumn{1}{c|}{0.9110/ 1.0245/ 1.2163}    &1.8466/ 2.0912/ 2.8277          & \multicolumn{1}{c|}{1.3746/ 1.4425/ 1.5039}    &4.0153/ 4.8510/ 5.3264          \\ \hline
GraphSage              & \multicolumn{1}{c|}{0.1196/ 0.1345/ 0.1708}    &0.5796/ 0.6047/ 0.6346           & \multicolumn{1}{c|}{0.9102/ 1.0530/ 1.2903}    &1.8473/ 2.0977/ 2.7651          & \multicolumn{1}{c|}{1.3966/ 1.4577/ 1.5106}    &4.0801/ 4.9626/ 5.2976          \\ \hline
GAE                    & \multicolumn{1}{c|}{0.1103/ 0.1312/ 0.1607}    &0.5701/ 0.6013/ 0.6297          & \multicolumn{1}{c|}{0.9132/ 1.0118/ 1.2664  }    &1.8412/ 2.1632/ 2.7061          & \multicolumn{1}{c|}{1.3719/ 1.4307/ 1.5006}    &4.0337/ 4.9795/ 5.2080          \\ \hline
POI                    & \multicolumn{1}{c|}{0.0933/ 0.1204/ 0.1578}    &0.5578/ 0.5903/ 0.6198          & \multicolumn{1}{c|}{0.8892/ 0.9911/ 1.0362}    &1.8277/ 2.1333/ 2.5391          & \multicolumn{1}{c|}{1.3316/ 1.3646/ 1.4297}    &3.9872/ 3.9189/ 4.8996          \\ \hline
HDGE                   & \multicolumn{1}{c|}{0.0865/ 0.1203/ 0.1524}    &0.5502/ 0.5840/ 0.6134          & \multicolumn{1}{c|}{0.8667/ 0.9889/ 1.0243}    &1.8251/ 1.9357/ 2.5376          & \multicolumn{1}{c|}{1.2997/ 1.3572/ 1.4214}    &3.9846/ 4.4336/ 4.8316          \\ \hline
ZE-Mob                 & \multicolumn{1}{c|}{0.1002/ 0.1245/ 0.1576}    &0.5668/ 0.5932/ 0.6154          & \multicolumn{1}{c|}{0.8900/ 0.9982/ 1.0443}    &1.8359/ 2.1669/ 2.5718          & \multicolumn{1}{c|}{1.3314/ 1.4008/ 1.4707}    &4.0366/ 4.5656/ 4.9116          \\ \hline
MV-PN                  & \multicolumn{1}{c|}{0.0903/ 0.1235/ 0.1571}    &0.5502/ 0.5843/ 0.6132          & \multicolumn{1}{c|}{0.8886/ 0.9791/ 1.0490}    &1.8313/ 1.9775/ 2.5687          & \multicolumn{1}{c|}{1.3306/ 1.3874/ 1.4525}    &3.9530/ 4.5368/ 4.9061          \\ \hline
CGAL                  & \multicolumn{1}{c|}{0.1013/ 0.1267/ 0.1502}    &0.5682/ 0.5998/ 0.6197          & \multicolumn{1}{c|}{0.9097/ 1.0504/ 1.0910}    &1.8557/ 1.9845/ 2.5870          & \multicolumn{1}{c|}{1.3353/ 1.4209/ 1.4833}    &4.0671/ 4.5850/ 4.9606          \\ \hline
MVURE                  & \multicolumn{1}{c|}{0.0874/ 0.1196/ 0.1503}    &0.5405/ 0.5710/ 0.6013          & \multicolumn{1}{c|}{0.8699/ 0.9364/ 0.9502}    &1.8157/ 1.9751/ 2.4356          & \multicolumn{1}{c|}{1.3007/ 1.3266/ 1.3347}    &3.6715/ 4.2534/ 4.7200          \\ \hline
MGFN                   & \multicolumn{1}{c|}{0.0831/ 0.1145/ 0.1492}    &0.5385/ 0.5671/ 0.5917          & \multicolumn{1}{c|}{0.8783/ 0.9376/ 0.9771}    &1.8163/ 1.9907/ 2.4198          & \multicolumn{1}{c|}{1.3266/ 1.3497/ 1.3561 }    &3.7514/ 4.3200/ 4.7311          \\ \hline \hline
\model                 & \multicolumn{1}{c|}{\textbf{0.0665/ 0.0675/ 0.0684}}    &\textbf{0.4931/ 0.4956/ 0.4971}         & \multicolumn{1}{c|}{\textbf{0.8364/ 0.8767/ 0.8801}}    &\textbf{1.8145/ 1.8864/ 2.3510}          & \multicolumn{1}{c|}{\textbf{1.2871/ 1.3134/ 1.3280}}    &\textbf{3.6446/4.2278/4.6836}          \\ \hline
\end{tabular}
\end{table*}

\begin{table}[h]
\centering
\footnotesize
\setlength{\tabcolsep}{0.3pt}
\caption{Data Description of Experimented Datasets}
\Description{The table provides statistic details of experiment data.}
\vspace{-2mm}
\label{fig:data_sta}
\begin{tabular}{c|c|c}
\hline
Data        & \textbf{Description of Chicago Data}                                                                                        & \textbf{Description of NYC data} \\ 
\hline 
Census  &\begin{tabular}[c]{@{}c@{}}Boundaries of 234 regions split by\\ streets in a certain district, Chicago\end{tabular}         &\begin{tabular}[c]{@{}c@{}}Boundaries of 180 regions split\\ by streets in Manhattan, New York\end{tabular}                             \\ \hline 
Taxi trips & \begin{tabular}[c]{@{}c@{}}Total 386,272 taxi \\trips during a month\end{tabular}                  &\begin{tabular}[c]{@{}c@{}} Total 1,445,285 taxi \\trips during a month\end{tabular}                             \\ \hline
Crime data     &\begin{tabular}[c]{@{}c@{}}Total 321,876 \\crime records during 1 year\end{tabular}   &\begin{tabular}[c]{@{}c@{}}Total 108,575 crime\\ records during 1 year\end{tabular}                             \\ \hline
POI data       &\begin{tabular}[c]{@{}c@{}} Total 3,680,125 POI locations\\ of 130 categories \end{tabular}  &\begin{tabular}[c]{@{}c@{}}Total 20,569 POI locations\\ of 50 categories\end{tabular} 
\\ \hline 
House price       &\begin{tabular}[c]{@{}c@{}}Total 44,447 house price data\\
in a certain district, Chicago\end{tabular}  &\begin{tabular}[c]{@{}c@{}}Total 22,540 house price data\\
in Manhattan, New York\end{tabular} 
\\  \hline
\end{tabular}
\end{table}

\subsection{Hyperparameter Studies (RQ5)}
\label{sec:parameter_study}
\begin{figure}
\centering
\footnotesize
\begin{tabular}{c}
  \begin{minipage}{0.160\textwidth}
	\includegraphics[width=\textwidth]{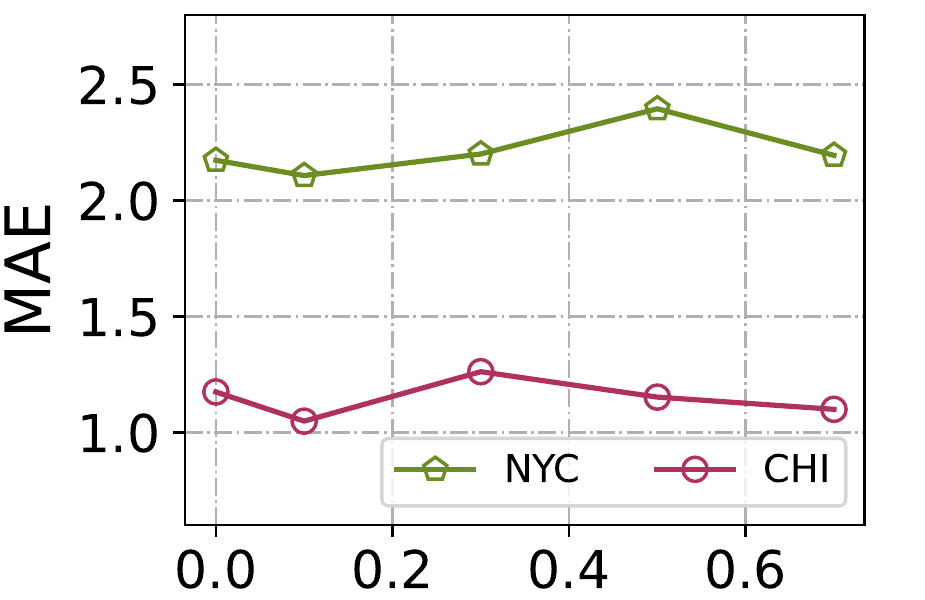}
  \end{minipage}\hspace{-1.5mm}
  
  


   \begin{minipage}{0.160\textwidth}
	\includegraphics[width=\textwidth]{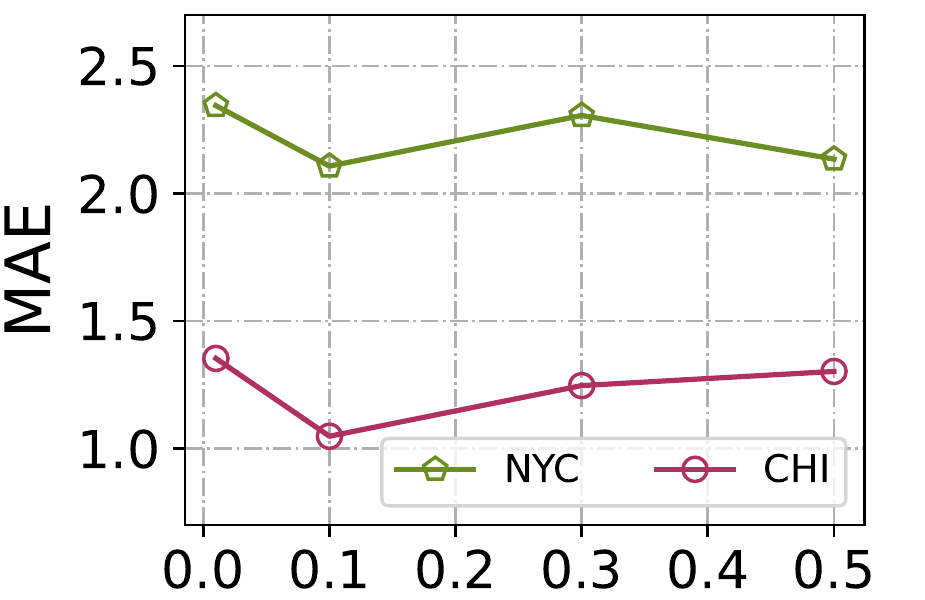}
  \end{minipage}\hspace{-1.5mm}


   \begin{minipage}{0.160\textwidth}
	\includegraphics[width=\textwidth]{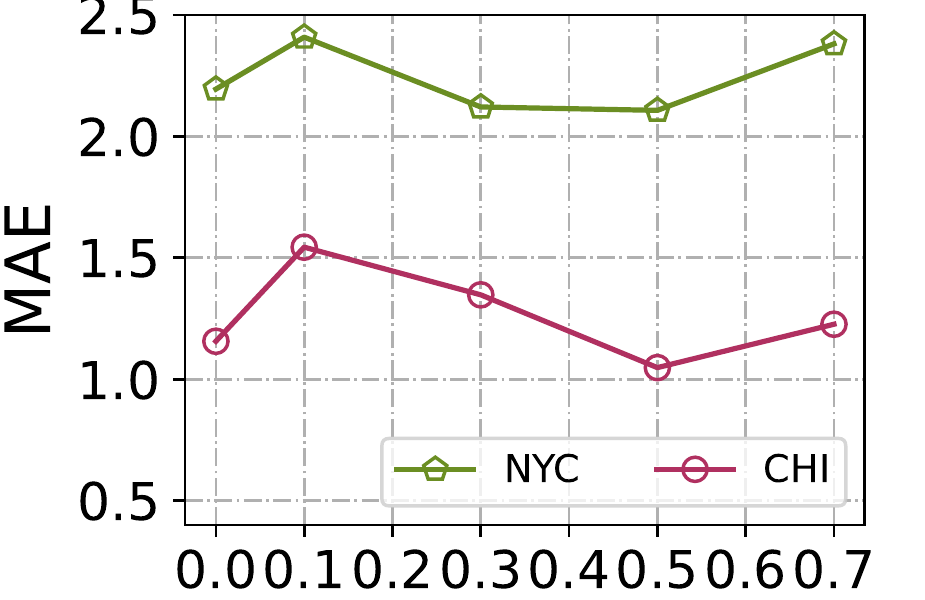}
  \end{minipage}\hspace{-1.5mm}

\\
\hspace{0.5mm}
 (a) MAE of $\beta$ 
 \hspace{12.5mm}
 (b) MAE of $\xi$
 \hspace{12.5mm}
 (c) MAE of $w_1$ 
\end{tabular}
\caption{Effect study for hyperparameters in the performance of \model\ on NYC crime data, in terms of MAE.}
\Description{The figure presents effects of different hyperparameters on final performance.}
\label{fig:param}
\end{figure}

To show the effect of different parameter settings, we conduct experiments to evaluate the performance of our framework \model\ with different configurations of important hyperparameters (\eg, (a) $\#$ $\beta$, $\#$ $\xi$ and $\#$ $w_1$)). When varying a specific hyperparameter for effect investigation, other parameters are fixed with default values. The results are shown in Figure~\ref{fig:param}. We summarize the observations below to analyze the influence of different hyperparameters:
\begin{itemize}[leftmargin=*]
    \item We vary $\beta$ from the range of $\left\{0.0, 0.1, 0.3, 0.5\right\}$. The best performance is achieved with $\beta = 0.1$. The prediction performance decreases as we further increase the value of $\beta$, which suggests larger $\beta$ does not always bring better model representation ability. The further increase of $\beta$ leads to little gap issue between positive samples and negative samples. \\\vspace{-0.12in}
    
    \item When $\xi$ is set as 0.1, the best performance can be achieved. Additionally, $w_1$ is the weight for the auxiliary supervised signal, which is utilized to calculate the distance between the positive samples and the negative samples. We observe that the best performance is achieved when $w_1 =0.5$. 
\end{itemize}

\begin{algorithm}[h]
    \caption{Automated Contrastive View Generation}
    \Description{We provide technical details of contrastive view generations.}
    \label{alg:viewGen}
    \KwIn{
        The multi-view spatio-temporal graph $\mathcal{G}$ consisting of three different graph layers;
    }
    \KwOut{
        Graph views $\mathcal{G}_1', \mathcal{G}_2'$ for contrastive learning;
    }
    Apply the cross-layer graph message passing on $\mathcal{G}$ for the low-dimensional graph representation $\textbf{H}$ (Eq~\ref{eq:multorder});\\   
    Employ the adjustable variational graph encoder for low-dimensional data generation $\tilde{\textbf{H}}$;\\
    Decode the hidden representations to graph structures and sparsify the adjacent matrix $\tilde{\textbf{P}}$;\\
    Use random walker to sample small sub-graph pairs $\mathcal{G}_1', \mathcal{G}_2'$ for the two generated graph views;\\
    \textbf{Return} $\mathcal{G}_1'$ and $\mathcal{G}_2'$.
\end{algorithm}

\begin{figure}
\centering
\begin{tabular}{c c }
\hspace{-25mm}
\begin{minipage}{0.5cm}
\includegraphics[width=7.5cm]{figures/title}
\end{minipage}\vspace{0.5mm}
&
\\\hspace{-3.0mm}
  \begin{minipage}{0.225\textwidth}
	\includegraphics[width=\textwidth]{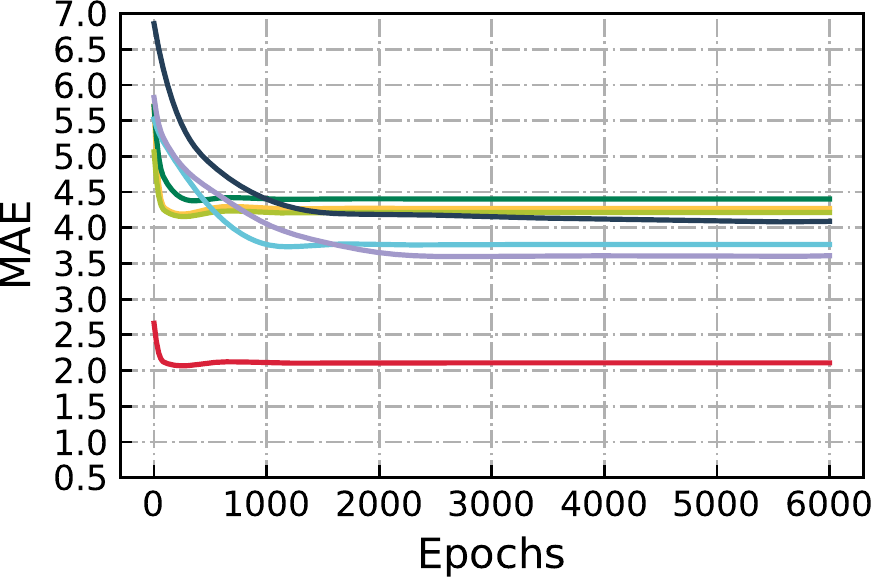}
  \end{minipage}\hspace{-3.0mm}
  &
  \begin{minipage}{0.225\textwidth}
    \includegraphics[width=\textwidth]{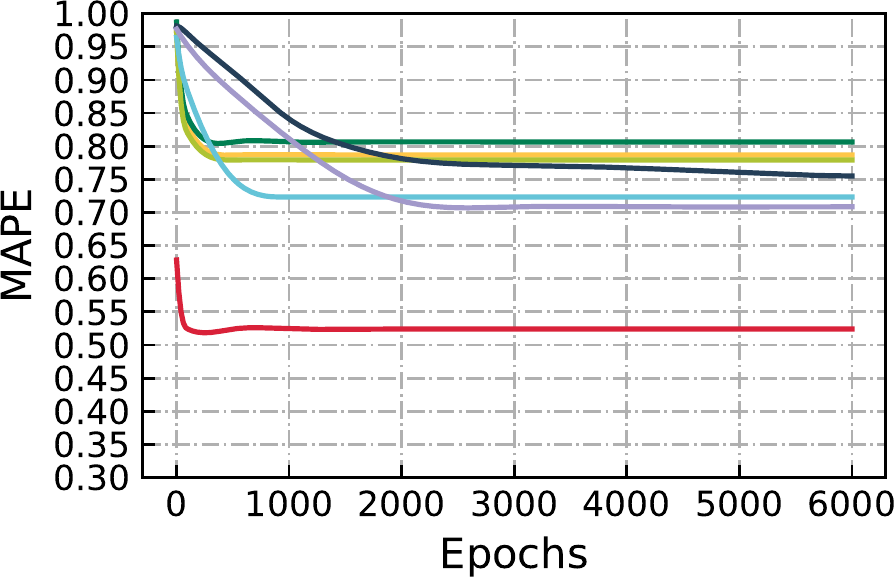}
  \end{minipage}\hspace{-4.0mm}

\end{tabular}
\caption{MAE and MAPE of crime prediction with epochs.}
\Description{We aim to provide performance with convergence speed of region representation methods.}
\label{fig:loss}
\end{figure}

\subsection{Model Convergence Study}
\label{sec:convergence}


To show our model convergence, Figure~\ref{fig:loss} presents the crime prediction performance using the encoded embeddings by different region representation methods \wrt\ epochs. From the results, we can find that our method \model\ converges faster than other region representation methods, meanwhile achieving the best performance. It validates the scalability of our \model\ method with our proposed automated spatio-temporal contrastive learning framework.


\begin{algorithm}[h]
    \caption{The \model\ Learning Algorithm}
    \Description{The algorithm presents the whole process of our method.}
    \label{alg:overall}
    \KwIn{
        The hierarchical spatio-temporal graph $\mathcal{G}$, the maximum epoch number $E$, the learning rate $\eta$;
    }
    \KwOut{
        Regional embeddings $\textbf{H}$
    }
    Initialize all parameters;\\
    \For{$e=1$ to $E$}{
        Generate contrastive views $\mathcal{G}_1', \mathcal{G}_2'$ following Alg~\ref{alg:viewGen};\\
        Use the hierarchical graph encoder for embeddings $\textbf{H}_1', \textbf{H}_2'$ \wrt~the generated views;\\
        Compute the CL loss $\mathcal{L}$ containing the InfoNCE $\mathcal{L}_\text{NCE}$ and the InfoBN $\mathcal{L}_\text{BN}$ (Eq~\ref{eq:overallLoss}, \ref{eq:bnLoss});\\
        Minimize $\mathcal{L}$ using Adam with learning rate $\eta$;\\
        Calculate the reward $\text{R}(\mathcal{G}, \theta_1, \theta_2)$ for view generation with InfoMin (Eq~\ref{eq:infomin_1});\\
        Calculate the reconstruction loss $\mathcal{L}_\text{Rec}$ of VGAE;\\
        Optimize the VGAE-based graph sampler by minimizing $\text{R}(\cdot)$ combining $\mathcal{L}_\text{Rec}$ (Eq~\ref{eq:opt_graphSamp});\\
    }
    \textbf{Return} the learned region embeddings $\textbf{H}$.
\end{algorithm}

\end{document}